\documentclass[lettersize,journal]{IEEEtran}
\usepackage{amsmath,amsfonts}
\usepackage{algorithmic}
\usepackage{algorithm}
\usepackage{array}
\usepackage[caption=false,font=normalsize,labelfont=sf,textfont=sf]{subfig}
\usepackage{textcomp}
\usepackage{stfloats}
\usepackage{url}
\usepackage{verbatim}
\usepackage{graphicx}
\usepackage[nocompress]{cite}
\usepackage{booktabs}
\usepackage{multicol}
\usepackage{threeparttable}
\usepackage{tabularx} 
\usepackage{arydshln} 
\usepackage{multirow} 
\usepackage[colorlinks=true, citecolor=blue, linkcolor=blue, urlcolor=black]{hyperref}

\begin{document}

\title{RailGen: Improving Railway Intrusion Detection via Agent-Guided Small-Scale Foreign Object Generation}

\author{Quan~Hao, Ziyang~Tao, Chenxi~Zhang, Yudong~Wang, Rui~Shi and~Liguo~Zhang,~\IEEEmembership{Senior Member,~IEEE}%
\thanks{Quan Hao, Chenxi Zhang, Rui Shi and Liguo Zhang are with the School of Information Science and Technology, Beijing University of Technology, Beijing, China. (email: haoquan@emails.bjut.edu.cn; zcx20041004@emails.bjut.edu.cn; ruishi@bjut.edu.cn; zhangliguo@bjut.edu.cn)}

\thanks{Ziyang Tao is with the College of Computer Science, Beijing University of Technology, Beijing, China. (email: taoziyang@emails.bjut.edu.cn;)}

\thanks{Yudong Wang is with the Institute of Automation, State Key Laboratory of Multimodal Artificial Intelligence Systems, Chinese Academy of Sciences, Beijing 100190, China. (email: wangyudong2018@ia.ac.cn)}%
\thanks{This work was supported by Beijing Natural Science Foundation (Grant No. L243026, 4244088), and National Natural Science Foundation of China (Grant No. U2233211, 62403017). \textit{(Corresponding author: Liguo Zhang.)}}}

\markboth{IEEE TRANSACTIONS ON MULTIMEDIA,~Vol.~27, May~2026}%
{Shell \MakeLowercase{\textit{et al.}}: A Sample Article Using IEEEtran.cls for IEEE Journals}


\maketitle

\begin{abstract}

Small-object detection under long-tailed data distributions is a fundamental yet challenging problem in multimedia. Railway Foreign Object Detection (RFOD) epitomizes this challenge with easily confused small intrusions and scarce samples.
To address these issues, we propose a generative-augmented detection paradigm that leverages multimodal image generation to enrich the feature space of rare, small objects.
We first construct RailGen, a multimodal image generation agent based on large models. Under semantic constraints, RailGen automatically invokes tools to generate railway scenes, calibrate intrusion positions, extract foreign objects, and fuse them into realistic intrusion effects. This process produces high-quality synthetic samples that effectively densify the feature representations of tail classes and complete the small-object feature space.
Within this paradigm, we further propose FocalDEIM, a detection framework designed to enhance training with generated data. FocalDEIM improves dense matching with Focal Modulation for better small object discrimination and adopts Focal Loss to emphasize hard samples, thereby alleviating blurred inter-class boundaries in complex railway scenes.
Experimental results demonstrate that RailGen can generate high-quality small-scale foreign objects, reducing the object pixel area by up to 58$\times$ and by 13.85$\times$ on average. Equipped with these challenging samples, our paradigm surpasses the baseline DEIM by 5.6\% and 7.5\% in $mAP_{50}$ and $mAP_{50\text{-}95}$ respectively, and outperforms existing state-of-the-art methods. Ablation studies verify RailGen’s feature-space enrichment and FocalDEIM’s boundary discrimination. The paradigm provides an effective multimodal generative solution for long-tailed small-object detection in safety-critical applications.

\end{abstract}


\section{Introduction}

\begin{figure*}[htbp]
     \centering
     \includegraphics[width=1.0\textwidth]{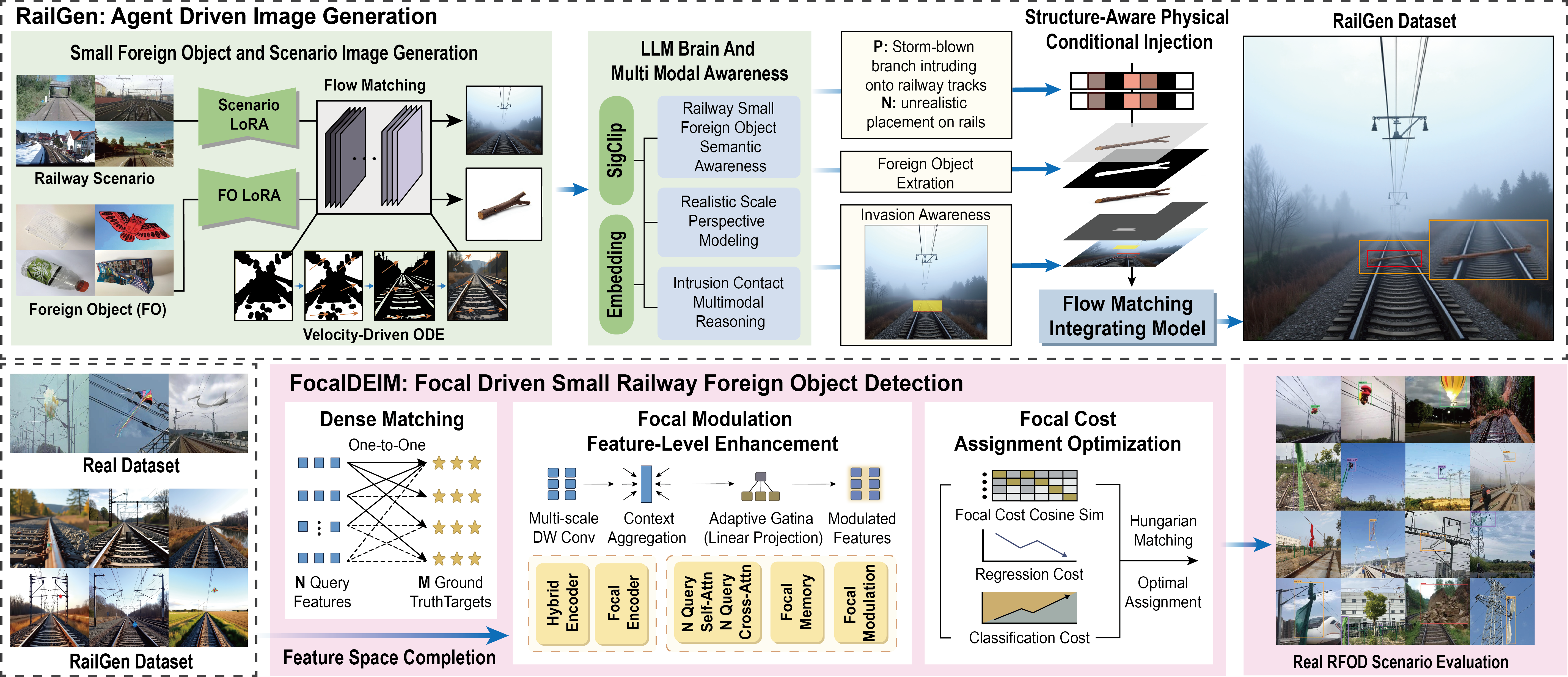}
     \caption{Overall pipeline of the generative-augmented detection paradigm for railway foreign object intrusion detection}
     \label{fig:framework}
\end{figure*}

\IEEEPARstart{S}{mall-object} detection under long-tailed data distributions is a fundamental yet challenging problem in multimedia, where rare categories with tiny object scales typically suffer from under-represented feature spaces and poor discrimination. Railway Foreign Object Detection (RFOD) epitomizes this challenge. In high-speed railway operation, even small artificial or natural objects, such as balloons, plastic films, stones, or tree branches, intruding at critical positions of the overhead contact line and rail tracks can cause serious accidents including facility damage, train derailment, and emergency stops. Accurate detection of such foreign objects is therefore essential for safe and stable high-speed railway operation~\cite{li2023yolov5s}.

However, high-speed railway operating environment is complicated. Lighting conditions change constantly, track textures are highly repetitive, and overhead line backgrounds easily cause visual confusion. These factors make it difficult for detection models to distinguish small-sized foreign objects from the background, hence degrading detection performance \cite{jin2025optimized}. Furthermore, real intrusion events are difficult to capture in practice. Existing datasets exhibit severe long-tail distributions, where normal scene samples vastly outnumber foreign object samples. This imbalance leads to an incomplete feature space, preventing detection models from learning discriminative representations of rare intrusion patterns.

Currently, railway foreign object intrusion detection primarily relies on object detection algorithms \cite{10719679, tian2025yolov12,lei2025yolov13}. In particular, DETR-series methods based on the Transformer architecture have demonstrated advantages in global modeling and object matching, while techniques such as Focal Modulation improve small object detection through feature enhancement \cite{huang2025deim,yang2022focal}. Nevertheless, these methods still rely heavily on limited real samples for training and thus continue to suffer from incomplete feature spaces and insufficient generalization to hard examples in long-tail distributions and complex background \cite{yang2022survey,alzubaidi2023survey}. Although image generation techniques have emerged as a promising avenue for alleviating sample scarcity \cite{pei2025neuro,10931847,rombach2022high,flux2024}, without accurate modeling of railway scenes and foreign object intrusion semantics, generative models are still prone to producing images with unreasonable object placement, scale distortion, and lighting inconsistency, making it difficult to complete the feature space and improve detection performance \cite{10902119,11175219,11237202}.

To address these issues, this paper proposes a generative-augmented detection paradigm for railway foreign object intrusion detection, where realistic image generation is leveraged to enrich the feature space for detection. 

Specifically, we first construct RailGen, a framework based on multimodal large models, to generate semantically consistent and physically plausible railway foreign object intrusion samples. First, with Low-Rank Adaptation (LoRA) for domain adaptation, Flow Matching is used to model continuous probability flows and generate railway background and foreign object images. Then, machine vision tools semantically analyze foreign object intrusion relationship and determine intrusion positions and scales that satisfy scene priors and physical constraints, thereby enabling the effective generation of challenging small-scale foreign objects in railway scenes. Furthermore, foreign object regional features are extracted and fused into the railway scene, producing high-fidelity intrusion images. This process increases sample diversity and helps complete the feature space under long-tail data distributions.

In addition to RailGen which enriches the feature space by expanding the sample distribution with diverse and physically plausible generated samples, we further propose FocalDEIM for joint training on generated and real data. Based on the dense one-to-one matching paradigm, FocalDEIM introduces Focal Modulation to refine decoder query representations before Hungarian assignment through context-aware aggregation, thereby enhancing the discriminability of small foreign object features in complex railway scenes. In addition, Focal Loss is adopted as the classification objective to reweight hard and easily confused samples during training, guiding optimization toward more informative matches and further improving detection performance.

Extensive experiments demonstrate that RailGen can generate high-quality small-scale foreign objects, reducing the object pixel area by up to $58\times$ and by $13.85\times$ on average. By leveraging these challenging samples, the proposed method improves $\text{mAP}_{50}$ and $\text{mAP}_{50\text{-}95}$ over the baseline model by $5.6\%$ and $7.5\%$ respectively, and outperforms existing state-of-the-art methods. Ablation studies further highlight the effectiveness of the proposed generation-detection synergy.

The main contributions of this paper are summarized as follows:
\begin{itemize}
\item \textbf{We propose RailGen}, a semantically consistent multimodal agent that generates realistic railway images through state-of-the-art generation performance, completing the feature space of detection models.

\item \textbf{We introduce FocalDEIM}, a detection framework that enhances generated-data training via Focal Modulation for small object discrimination in dense matching and Focal Loss for hard-sample learning.

\end{itemize}

\section{Related Work}
\subsection{Railway Foreign Object Detection}

In railway foreign object detection (RFOD), recent studies have demonstrated the feasibility of automatic detection using surveillance videos or images \cite{10719679,bai2026rosd,cao2022effective}. Nevertheless, existing methods still face two fundamental challenges: severe data scarcity leads to an incomplete feature space, and complex railway environments blur inter-class feature boundaries.

Transformer-based DETR performs global contextual modeling through encoder-decoder structures and self-attention, showing advantages in complex scene understanding \cite{zhao2024detrs,shehzadi2025object}. However, its one-to-one (O2O) matching mechanism suffers from sparse supervision, which further worsens the feature-space incompleteness when positive samples are scarce. To deal with this issue, DEIM replaces conventional label assignment with dense matching between predictions and ground-truth boxes, and enhances sensitivity to hard examples by introducing learnable matching costs \cite{huang2025deim,dai2023adaptive}. Built upon this paradigm, recent studies combine DEIM with multi-scale feature modulation to improve localization and recognition under complex background while maintaining inference efficiency \cite{deng2021extended,yin2024sdpdet}. In particular, Focal Modulation adaptively aggregates multi-scale contextual information through gated interactions, which benefits the discriminative representation of small objects and helps clarify ambiguous inter-class boundaries \cite{yang2022focal,cui2026focal}. Despite these advances, detection-centric methods remain constrained by an incomplete training feature space. Without sufficient diverse samples covering rare intrusion patterns, the boundary between foreign objects and background clutter remains poorly defined, leading to limited generalization in extreme operating conditions.

Moreover, RFOD is limited by the few-sample problem \cite{11237202}. Although generated samples have been introduced to augment training data \cite{zhang2024railway}, the domain gap between synthetic and real images restricts the resulting performance gains. Existing generation methods still fail to produce physically consistent and semantically valid samples that can effectively expand the feature space. As a result, they cannot adequately resolve the issue of the inter-class ambiguity between small foreign objects and complex railway background.

\subsection{Image Generation}

To address feature-space incompleteness caused by data scarcity, generative augmentation has emerged as a promising direction. From VAE \cite{kingma2022autoencodingvariationalbayes} and GAN \cite{goodfellow2014generativeadversarialnetworks} to recent diffusion-based models, especially Stable Diffusion \cite{rombach2022high}, image generation has achieved substantial progress in visual fidelity and training stability. More recently, Flow Matching \cite{lipman2023flowmatchinggenerativemodeling} has been introduced into image generation. FLUX adopts Rectified Flow to construct nearly linear deterministic ODE trajectories, enabling improved controllability with fewer sampling steps \cite{labs2025flux1kontextflowmatching,flux2024}. Combined with Low-Rank Adaptation (LoRA), these models can adapt to target domains with limited data by training only lightweight low-rank adapters, thereby improving the visual consistency of generated samples with railway scenes \cite{hu2022lora}.

However, existing generative models still struggle to achieve controllable railway foreign object intrusion generation. In particular, they lack explicit control over the intrusion semantics, including object category, placement, scale, and scene compatibility. This limitation becomes more severe when it comes to small-scale foreign objects, the visual cues of which are inherently weak and highly sensitive to scale distortion and background interference. As a result, directly generated intrusion images often suffer from unreasonable object placement, inaccurate object scale, and weak semantic alignment with the scene. These issues not only reduce physical plausibility, but also limit the effectiveness of generated samples in expanding the discriminative feature space, especially for hard small-object cases.

Recent advances in multimodal foundation models and multimodal agents provide new opportunities to address this limitation. Multimodal agents built on models such as Gemma exhibit strong visual understanding, task planning, and instruction-following capabilities, enabling them to decompose high-level objectives into executable steps \cite{gemmateam2025gemma3technicalreport,yin2025foodlmm,wang2025detailed}. Meanwhile, progress in vision-language models and image segmentation methods, including CLIP, CLIPSeg, and SAM, offers effective tools for object-aware parsing and stepwise visual processing \cite{radford2021learningtransferablevisualmodels,lueddecke22_cvpr,kirillov2023segment}. Furthermore, multimodal reasoning enhances the ability of agents to guide generation and evaluate semantic consistency \cite{gu2024survey,zhou2026semantic}. Nevertheless, existing studies have not yet established a multimodal agent framework that may bridge high-level semantic reasoning with fine-grained generation control for controllable RFOD synthesis, particularly for small-scale foreign object intrusion generation under physical and semantic constraints.

\section{Methods}

Data scarcity in railway scenarios leads to an incomplete feature space, blurred inter-class boundaries, and unstable matching in small object detection. To address these issues, this paper proposes a generation-detection collaborative framework. RailGen completes the feature space of railway foreign objects through synthetic data generation. FocalDEIM serves as a detector that is more sensitive to small railway foreign object representations and the expanded feature space, improving small object detection performance.

\subsection{RailGen: Model-Agent Collaborative Generation for Feature Space Completion}
\setlength{\parindent}{2em}

Existing end-to-end generation methods treat the process of image generation as black-box pixel regression. They do not explicitly model physical rules in railway scenarios (such as perspective, gravity, and safety distances) or foreign object semantic attributes (such as material, weight, and motion patterns). As a result, the generated samples often exhibit physical artifacts, such as floating objects, scale distortion, and inconsistent lighting. To address these issues, this paper presents RailGen, an agent-based generation framework built on multimodal large models. RailGen uses specialized tools across multiple stages to generate high-quality high-speed railway (HSR) foreign object intrusion images. As shown in Fig.~\ref{fig:framework}, these tools support intrusion localization, image generation, segmentation, and foreign object fusion.

Specifically, RailGen formulates generation as a physically constrained and semantically controllable optimization process via a perception–reasoning–action closed loop. The agent first predicts physically reasonable intrusion positions through visual-language semantic understanding and then allocates Flow Matching generation and semantic-aware fusion modules to maintain geometric alignment, lighting consistency, and occlusion relations. Through this multi-stage pipeline, RailGen synthesizes visually realistic and physically plausible samples, enriching the feature space for HSR foreign object detection.

\subsubsection{Multimodal Semantic Reasoning and Anchor Region Calibration}

Existing generation methods mostly adopt random or heuristic position placement. They ignore perspective constraints and physical common sense in railway scenes, resulting in foreign object positions that violate geometric consistency. To address this, we define anchor regions as candidate spatial locations satisfying physical plausibility constraints (e.g., heavy objects contact the trackbed, lightweight objects may be suspended), and calibrate these regions via multimodal semantic reasoning to enforce geometric consistency.

As shown in Fig.~\ref{fig:framework}, given a railway background image \( I_{\text{bg}} \in \mathbb{R}^{H \times W \times 3} \) and a description of the foreign object \( T \), we first extract joint multimodal representations through the SigLIP-ViT encoder, yielding a unified embedding vector. Rather than performing simple coordinate regression, we formulate the intrusion anchor point localization as a constrained spatial optimization problem. The search space is the set of candidate bounding boxes, where each candidate box is represented as a 4‑tuple of normalized center coordinates and width/height parameters.

The optimal anchor region \( b^* \) is obtained by maximizing a joint objective function of visual naturalness and railway physical consistency:
\begin{equation}
b^* = \arg\max_b \left( R(b \mid I_{\text{bg}}) + \lambda D(b \mid I_{\text{bg}}, \Theta_{\text{rail}}) \right).
\end{equation}
Here, \( R(\cdot) \) measures the visual consistency between the candidate region and the background's semantic features, implemented as a learnable compatibility function based on the backbone network's features. The second term, \( D(\cdot) \), encodes physical constraints related to railway scene, with parameters \( \Theta_{\text{rail}} \) including track geometry priors, gravity consistency, and key safety contact regions.

The physical constraint term can be decomposed as:
\begin{equation}
D(b \mid I_{\text{bg}}, \Theta_{\text{rail}}) = \alpha_1 \mathcal{C}_{\text{contact}}(b) + \alpha_2 \mathcal{C}_{\text{gravity}}(b) + \alpha_3 \mathcal{C}_{\text{persp}}(b),
\end{equation}
where \( \mathcal{C}_{\text{contact}}(b) \) enforces the physical contact relationship between the foreign object and the trackbed or contact network (distinguishing between heavy and lightweight objects based on mass attributes);  
\( \mathcal{C}_{\text{gravity}}(b) \) penalizes spatial configurations that violate gravity direction consistency;  
\( \mathcal{C}_{\text{persp}}(b) \) ensures that the scale and track perspective geometry are consistent.  

This modeling approach transforms railway safety knowledge and physical priors into computable spatial constraints, preventing the generation of physically unreasonable anchor positions and thus avoiding feature space contamination.

\subsubsection{Deterministic Generation Based on Flow Matching}

Traditional diffusion models construct non-deterministic curved sampling trajectories through Stochastic Differential Equations (SDEs), making precise control difficult. They also exhibit poor adaptability to structured railway scenes. To address this, we propose a Flow Matching generation paradigm with Low-Rank Adaptation that achieves cross-domain semantic alignment through deterministic probability paths.

We define a probability path $p_t$ from the noise distribution $p_1$ to the data distribution $p_0$, governed by the velocity field $v_t$ via a probability flow ODE. For training samples $x_0$ and Gaussian noise $\epsilon$, we use linear interpolation and optimize the Flow Matching loss.

During inference, the ODE is solved by integrating from $x_1$ to $x_0$:
\begin{equation}
    x_0 = x_1 + \int_{1}^{0} v_\theta(x_t, t, c) \, dt,
\end{equation}

\noindent iteratively solved using a numerical integrator at discrete timesteps $\{t_i\}_{i=0}^K$.

To achieve railway domain adaptation, the pretrained velocity field $v_\theta$ is frozen, and low-rank decomposition $\Delta W = BA$ is introduced for fine-tuning:
\begin{equation}
    v_\theta'(x_t, t, c) = v_\theta(x_t, t, c) + W_{\text{out}}(BA \cdot \text{proj}(x_t, t, c)).
\end{equation}

Intuitively, Flow Matching enables deterministic and controllable image synthesis by learning a direct transformation from noise to data along a linearized probability path. This property is particularly suitable for railway scenarios, where geometric consistency and precise spatial control are critical.

This strategy achieves a precise mapping from general visual distribution to railway scene manifold. After generating the HSR scene and foreign object images, we perform image segmentation on the foreign object images, providing geometric priors for subsequent fusion.

\subsubsection{Structure-Aware Physical Conditional Injection Mechanism}

Traditional image fusion methods (e.g., $\alpha$-blending and Poisson fusion) treat the foreground and background as independent pixel sets for post hoc composition, neglecting scene geometry consistency and physical constraints. This often leads to boundary artifacts, scale inconsistency, and lighting conflicts. Motivated by the conditional guidance idea of ControlNet \cite{zhang2023adding}, we propose a Structure-Aware Physical Conditional Injection (SPCI) mechanism, as shown in Fig.~\ref{fig:framework}. SPCI embeds fusion into the Flow Matching generation trajectory and achieves progressive, physically consistent composition through latent-space feature modulation.

Given the inferred foreign object configuration, we construct a Structure-aware Physical Conditional Injection (SPCI) representation:
\begin{equation}
\mathbf{S} = \{S_k\}_{k=1}^{5}, \quad S_k \in \mathbb{R}^{H \times W},
\end{equation}
where the five channels correspond to $S_{\text{mask}}$, $S_{\text{contour}}$, $S_{\text{depth}}$, $S_{\text{support}}$, and $S_{\text{illum}}$, encoding the foreground mask, contour, depth approximation, structural support, and lighting direction respectively.

Instead of heuristic gravity constraints, we model steady-state physical consistency. The support constraint is defined as:
\begin{equation}
\begin{aligned}
S_{\text{support}}(x,y)
&= \pi_{\text{heavy}}(T)\cdot
\mathbb{I}\big(
\mathrm{dist}((x,y), \mathcal{S}_{\text{ground}}) < \epsilon_g
\big) \
\\
&\quad + \pi_{\text{hang}}(T)\cdot
\mathbb{I}\big(
\mathrm{dist}((x,y), \mathcal{S}_{\text{wire}}) < \epsilon_w
\big),
\end{aligned}
\end{equation}

\noindent where $\mathcal{S}_{\text{ground}}$ denotes weight-bearing structures (e.g., trackbed and ballast),
$\mathcal{S}_{\text{wire}}$ denotes suspension structures (e.g., contact wires),
and $\pi_{\text{heavy}}(T)$ and $\pi_{\text{hang}}(T)$ denote
 object-dependent weights inferred from the semantic description $T$.

This formulation enforces physically plausible support conditions: heavy objects require structural support, while lightweight objects (e.g., balloons or kites) may remain suspended.

Additional channels further enforce perspective-depth consistency and lighting coherence, ensuring that object scale aligns with track perspective and that shading is consistent with the global light source.

We inject the SPCI tensor into latent features via a structure-aware modulation operator:
\begin{equation}
\mathbf{h}_{\text{fused}} = \mathbf{h} + \sum_{k=1}^{K} \gamma_k^0 \, \Psi_k \big(S_k, \mathcal{G}_{\text{rail}} \big),
\end{equation}
where $\mathbf{h}$ denotes Flow Matching features, $\gamma_k^0$ are learnable weights, $\Psi_k(\cdot)$ encodes constraint-specific modulation, and $\mathcal{G}_{\text{rail}} = \{\mathcal{L}_{\text{track}}, \mathcal{H}_{\text{wire}}\}$ represents railway geometry priors.

At each ODE step, the velocity field is modulated as:
\begin{equation}
x_{t-1} = x_t + \Delta t \cdot v_\theta \big(x_t, t, c, \mathbf{h}_{\text{fused}}, \mathcal{G}_{\text{rail}} \big),
\end{equation}
jointly guided by semantic and structural constraints. The modulation weights are dynamically updated:
\begin{equation}
\gamma_k(t) = \gamma_k^0 \cdot \omega_{\text{depth}}(t),
\end{equation}
enabling progressive enforcement of physical consistency along the generation trajectory.

Overall, SPCI integrates geometric, structural, and physical constraints directly into the generative process, reducing artifacts and improving realism.

\subsection{FocalDEIM: Focal-Driven Dense Matching for Small Objects}
\label{sec:detection}

RailGen expands the feature space by providing diverse and physically plausible synthetic samples. Building on this enriched space, dense matching of DEIM increases positive supervision by assigning each ground-truth target to multiple spatially adjacent queries. However, this process exposes a new bottleneck: because small foreign objects share similar local patterns with track textures and the background, they still suffer from insufficient discriminability during Hungarian matching, even with richer candidates.

To address this, we propose FocalDEIM, as illustrated in Fig.~\ref{fig:framework}. We first introduce Focal Modulation to enhance query feature representations via context-aware aggregation, improving the discriminability of small object features before assignment. We then incorporate a focal-aware matching cost into the Hungarian objective, which measures semantic consistency between queries and targets and guides the model toward more effective learning and faster convergence.

\subsubsection{Dense Matching in the RailGen-Enriched Feature Space}

DEIM follows the DETR end-to-end set prediction paradigm, extracting image features through Transformer encoders, then converting $N$ learnable query vectors $\mathbf{Q} \in \mathbb{R}^{N \times C}$ into class predictions $\hat{Y}_{\text{cls}} \in \mathbb{R}^{N \times K}$ (where $K$ is the number of classes) and bounding box coordinates $\hat{Y}_{\text{box}} \in \mathbb{R}^{N \times 4}$ via the decoder.

Built upon the enriched feature space provided by RailGen, the model performs label assignment between predictions and ground-truth targets. During training, $N$ prediction results must be optimally assigned to $M$ targets ($N \gg M$), which is achieved through Hungarian Matching. Define the matching cost matrix $\mathbf{C} \in \mathbb{R}^{M \times N}$, whose elements comprise classification and regression costs:
\begin{equation}
\begin{aligned}
\mathcal{C}_{\text{match}}(i,j) =\; & -\hat{p}_{\sigma(i)}(c_i) \\
& + \lambda_{\text{L1}}\|\mathbf{b}_i - \hat{\mathbf{b}}_{\sigma(j)}\|_1 \\
& + \lambda_{\text{GIoU}}\mathcal{L}_{\text{GIoU}}(\mathbf{b}_i, \hat{\mathbf{b}}_{\sigma(j)}),
\end{aligned}
\end{equation}

\noindent where $\hat{p}(c_i)$ denotes the predicted probability for class $c_i$, and $\mathbf{b}$ represents bounding box parameters. The optimal assignment $\hat{\sigma}$ is obtained by solving:
\begin{equation}
    \hat{\sigma} = \operatorname*{arg\,min}_{\sigma \in \mathfrak{S}_N} \sum_{i}^M \mathcal{C}_{\text{match}}(i, \sigma(i)).
\end{equation}

While the enriched feature space provides more complete visual coverage of railway foreign objects, the supervision signal remains sparse under the DETR-style one-to-one matching paradigm. To better exploit this feature space, DEIM introduces Dense One-to-One Matching by applying $S$ different scales and cropping augmentations $\{\mathcal{A}_s\}_{s=1}^S$ to training images. This allows each ground-truth target to be matched with multiple spatially adjacent query vectors, expanding the number of positive samples by a factor of $S$ and alleviating the sparse supervision problem.

However, even with the enriched feature space, small foreign objects still exhibit weak feature responses and remain highly similar to background textures. During early training stages, randomly initialized query vectors produce decoder features with limited discriminability, making small objects easily confused with the background in Hungarian matching. As a result, even with abundant candidates provided by dense matching, accurate assignment remains difficult to guarantee due to insufficient discriminability in the feature space.

\subsubsection{Context-Aware Matching Based on Focal Modulation}

Specifically, we propose FocalBlock as a context-aware feature enhancement module that refines decoder query representations before Hungarian assignment, improving small object discriminability against cluttered background. We further introduce FocalLoss as the classification objective to reweight hard and easily confused samples during training, stabilizing optimization and accelerating convergence toward correct assignments.

To address the insufficient discriminability of small objects during Hungarian matching, we introduce Focal Modulation mechanism into matching cost computation. This mechanism enhances small object features through explicit context aggregation and feature enhancement. Given decoder output query feature sequences $\mathbf{F}_{\text{query}} \in \mathbb{R}^{N \times C}$, we first aggregate context information using multi-scale depthwise separable convolutions:
\begin{equation}
    \mathbf{F}_{\text{ctx}}^{(k)} = \text{DWConv1D}_k(\mathbf{F}_{\text{query}}), \quad k=1,\dots,K,
\end{equation}
\begin{equation}
    \mathbf{G}_k = \text{Softmax}_k(\mathbf{W}_g \mathbf{F}_{\text{query}}), \quad \mathbf{F}_{\text{agg}} = \sum_{k=1}^{K} \mathbf{G}_k \odot \sigma(\mathbf{F}_{\text{ctx}}^{(k)}),
\end{equation}

\noindent where $\mathbf{G}_k$ represents adaptive gating weights with linear projection (no activation). The aggregated context is then injected into the original queries through projection layers, yielding modulated features:
\begin{equation}
    \mathbf{F}_{\text{mod}} = \mathbf{F}_{\text{query}} + \mathbf{W}_p \mathbf{F}_{\text{agg}}.
\end{equation}

We define the focal matching cost based on cosine similarity, measuring the semantic consistency between modulated query features and ground-truth target features $\mathbf{F}_{\text{tgt}} \in \mathbb{R}^{M \times C}$ (obtained by selecting the query with the highest predicted classification score for each ground-truth target from the current decoder output):
\begin{equation}
    \mathcal{C}_{\text{focal}}(i,j) = 1 - \frac{\mathbf{F}_{\text{mod}}^{(j)} \cdot \mathbf{F}_{\text{tgt}}^{(i)\top}}{\|\mathbf{F}_{\text{mod}}^{(j)}\| \|\mathbf{F}_{\text{tgt}}^{(i)}\|}.
\end{equation}

The final matching cost matrix is composed of weighted classification, regression, and focal modulation costs:
\begin{equation}
    \mathcal{C}_{\text{total}} = \lambda_{\text{cls}}\mathcal{C}_{\text{cls}} + \lambda_{\text{box}}\mathcal{C}_{\text{box}} + \lambda_{\text{focal}}\mathcal{C}_{\text{focal}}.
\end{equation}

\textbf{Training Objective.} For the matched positive sample set $\mathcal{P}$ and unmatched negative samples, the total training loss is defined as:
\begin{equation}
\begin{aligned}
\mathcal{L}_{\text{total}} =
& \sum_{i \in \mathcal{P}}
\left[
\mathcal{L}_{\text{cls}}(c_i, \hat{c}_{\hat{\sigma}(i)})
+ \mathcal{L}_{\text{box}}(\mathbf{b}_i, \hat{\mathbf{b}}_{\hat{\sigma}(i)})
\right] \\
& + \lambda_{\text{neg}}
\sum_{j \notin \mathcal{P}}
\mathcal{L}_{\text{cls}}(\emptyset, \hat{c}_j),
\end{aligned}
\end{equation}

\noindent
where $\mathcal{L}_{\text{cls}}$ denotes the Focal Loss, which alleviates class imbalance, and $\mathcal{L}_{\text{box}}$ represents the combination of GIoU Loss and L1 Loss.

Through the above mechanism, FocalDEIM achieves effective learning of small foreign objects with limited training samples. The dense matching strategy of DEIM alleviates sparse supervision caused by data scarcity by expanding the positive sample candidate space. The Focal Modulation mechanism ensures accurate identification of these candidates during Hungarian matching through explicit enhancement of small object feature representations, resolving the problem of insufficient discriminability caused by blurred inter-class boundaries. This module operates only during label assignment in the training stage and is completely removed during inference, improving small object detection performance with zero additional computational overhead.

\section{Results}
\subsection{Experimental Setup}
\subsubsection{Datasets and Preprocessing}
We use six datasets in our experiments:

\textbf{(1) Source Dataset.} containing 4,000 real railway scene images and 4,131 foreign-object images for training the RailGen image generation model.
\textbf{(2) RailGen Dataset.} containing 1,318 RFOD images generated by RailGen and manually annotated for label accuracy. Empirically, augmenting with 400 generated samples gives the best detection performance.
\textbf{(3) Real Train Set.} containing 398 real RFOD images with manually annotated small foreign objects for training.
\textbf{(4) Real Validation Set.} containing 102 real RFOD images with manually annotated small foreign objects for validation and evaluation.
\textbf{(5) CES (Catenary Electrical System Dataset).} Used as an auxiliary augmentation source to assess cross-domain generalization.
\textbf{(6) RailFOD23}~\cite{Chen2024RailFOD23}. A public benchmark dataset used for comparison.

\subsubsection{Evaluation Metrics for Image Generation}

These datasets support image generation, detector training, and benchmarking. Following the LLM-as-a-Judge paradigm~\cite{GU2026101253}, we use Gemini-3-Pro to evaluate generated railway foreign-object images using three metrics: \textbf{Scene Realism (SR)}, \textbf{Foreign Object Visual Quality (FOVQ)}, and \textbf{Foreign Object Plausibility (FOP)}, each scored from 0 to 10. The average score is computed as
\[
\textbf{Avg} = \frac{\textbf{SR} + \textbf{FOVQ} + \textbf{FOP}}{3}.
\]
The evaluator also provides strengths, problems, and suggestions, with judgments grounded in real-world physics.

To further assess small-object generation, we measure foreign-object pixel area. \textbf{FO Pixel} denotes the mean object pixel count, while \textbf{Avg Ratio} and \textbf{Max Ratio} denote the average and maximum pixel-count ratios relative to the reference method. These metrics reflect the ability to generate small but recognizable foreign objects.

\begin{figure}[htbp!]
    \centering
    \includegraphics[width=0.48\textwidth]{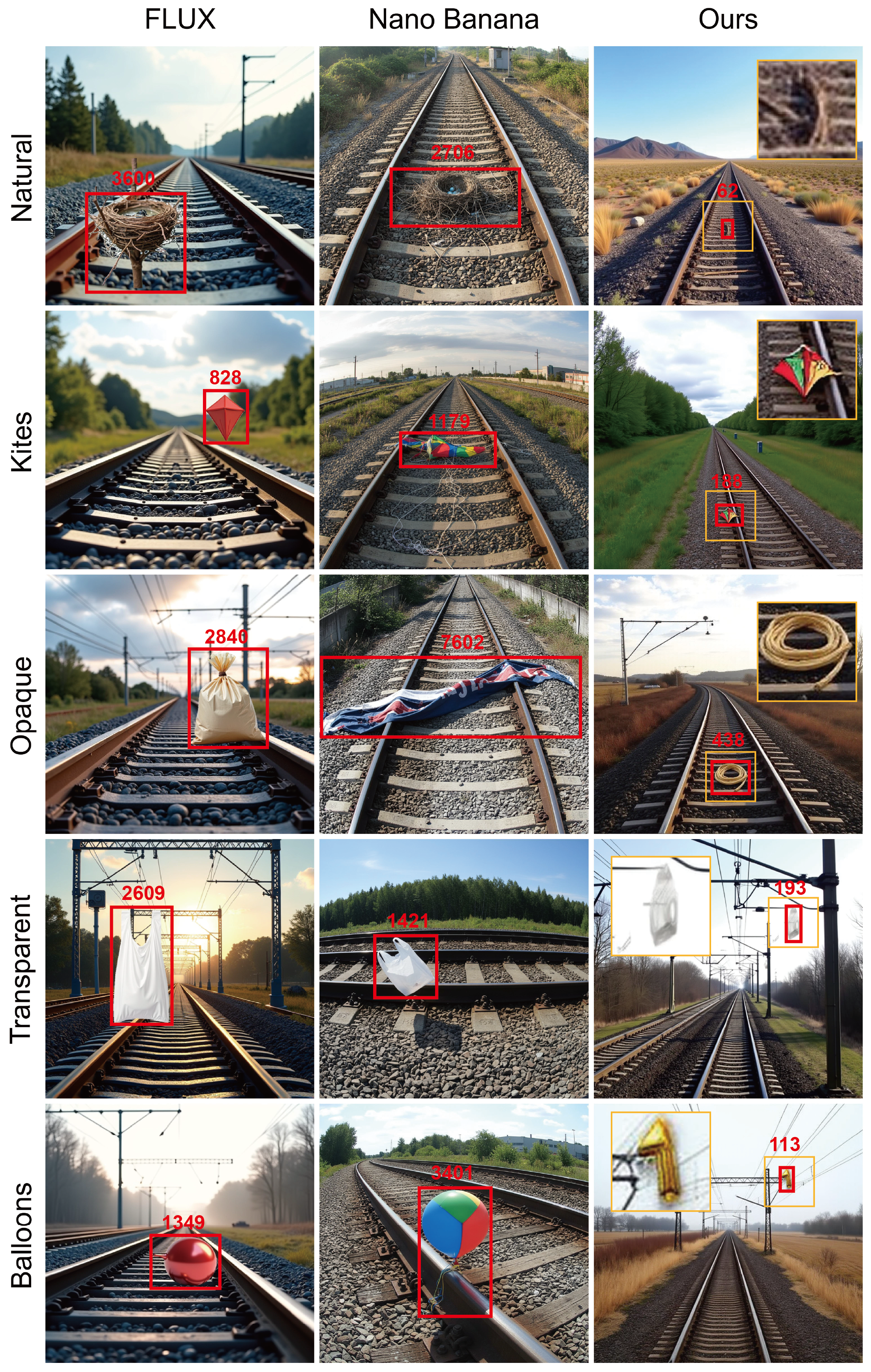}
    \caption{Comparison of state-of-the-art (SOTA) open-source and closed-source image generation models. Red boxes indicate intruding foreign objects, and red numbers denote their pixel counts. Green boxes show enlarged views of the small foreign objects generated by our method.}
    \label{fig:compare}
\end{figure}

\begin{table}[htbp!]
\centering
\caption{Comparison of generation results with SOTA methods}
\label{tab:generative_models}
\begin{threeparttable}
\begin{tabularx}{\columnwidth}{l XXX}
\toprule
Metric & FLUX~\cite{flux2024} & NanoBanana~\cite{nanobanana} & \textbf{RailGen (Ours)} \\
\midrule
\textbf{SR}      & 5.83  & 6.02  & \textbf{6.42} \\
\textbf{FOVQ}     & 2.66  & 4.30  & \textbf{4.95} \\
\textbf{FOP}     & 1.13  & 3.11  & \textbf{4.26} \\
\textbf{Avg} & 3.21  & 4.48 & \textbf{5.21} \\
\midrule
\textbf{FO Pixel}      & 2245.2 & 3261.8 & \textbf{198.8} \\
\textbf{Avg Ratio}    & 11.3$\times$ & 16.4$\times$ & \textbf{1.0$\times$} \\
\textbf{Max Ratio}    & 58.0$\times$ & 43.6$\times$ & \textbf{1.0$\times$} \\
\bottomrule
\end{tabularx}
\end{threeparttable}
\end{table}

To systematically evaluate the effect of different generative models on data quality, we conduct an automated assessment using Gemini-3-Pro as the evaluation agent. The compared methods include (1) FLUX~\cite{flux2024}, a SOTA open-source method; (2) Nano Banana~\cite{nanobanana}, a SOTA closed-source method; and (3) RailGen, our proposed method. Table~\ref{tab:generative_models} reports the evaluation scores, while Fig.~\ref{fig:compare} presents representative visual comparisons. Notably, the plastic bags generated by our method naturally hang on utility poles, demonstrating strong physical consistency.

As shown in Table~\ref{tab:generative_models}, our method achieves the highest scores in SR, FOVQ, FOP, and the Avg metric, indicating superior generation quality over existing methods. Nano Banana ranks second, while FLUX performs the worst.

Beyond perceptual quality, we further analyze foreign object scale. Existing methods generally produce much larger objects. Under the same $768\times768$ resolution, FLUX and Nano Banana generate 2245.2-pixel objects and 3261.8-pixel objects respectively on average, while our method generates 198.8-pixel objects on average. In other words, their generated objects are 11.3$\times$ and 16.4$\times$ larger than ours on average, with maximum ratios of 58.0$\times$ and 43.6$\times$ respectively. These results show that our method can generate much smaller yet still recognizable and physically plausible foreign objects, thereby introducing more  hard-to-detect samples for downstream detection.

\subsection{Comparison of Generation Quality with Existing Datasets}

\begin{table}[htbp!]
\centering
\caption{Comparison of generation quality with existing datasets}
\label{tab:reference}
\small
\begin{threeparttable}

\setlength{\tabcolsep}{4pt}

\begin{tabularx}{\columnwidth}{l XXXX}
\toprule
& \textbf{SR} 
& \textbf{FOVQ} 
& \textbf{FOP} 
& \textbf{Avg}  \\
\midrule
SODA10M \cite{han2021soda10m}            & 0.00 & 0.17 & 0.01 & 0.06 \\
STL-10 \cite{coates2011analysis}             & 0.02 & 2.04 & 0.05 & 0.70 \\
CES              & 2.17 & 3.96 & 4.03 & 3.39 \\
RailFOD23 \cite{Chen2024RailFOD23}          & \textbf{6.66} & 2.64 & 1.33 & 3.54 \\
\textbf{RailGen (Ours) }      & 6.42 & \textbf{4.95} & \textbf{4.26} & \textbf{5.21} \\
\hdashline
Real RFOD Scenes    & 5.59 & 7.14 & 7.80 & 6.84 \\
\bottomrule
\end{tabularx}

\end{threeparttable}
\end{table}

\begin{figure}[htbp]
    \centering
    \includegraphics[width=0.85\linewidth]{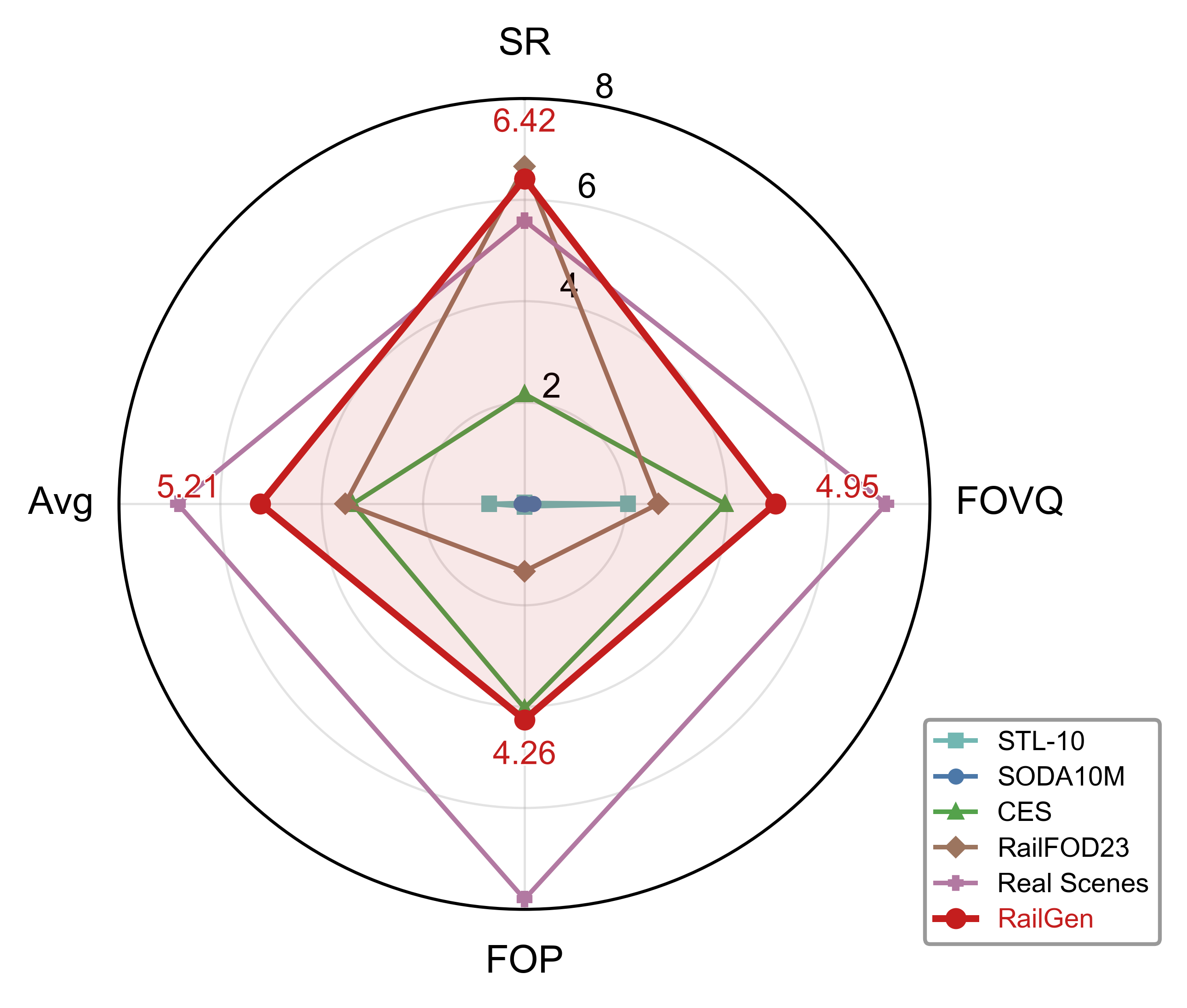}
    \caption{Radar-chart visualization of comparisons in generation quality.}
    \label{fig:radar_generation_comparison}
\end{figure}

Table~\ref{tab:reference} and Fig.~\ref{fig:radar_generation_comparison} present the evaluation results across different datasets. We include two unrelated datasets, SODA10M \cite{han2021soda10m} (a generic driving dataset) and STL-10 \cite{coates2011analysis} (a natural image classification dataset), as negative controls. Both datasets receive extremely low scores, showing that the evaluator can reliably distinguish irrelevant data from railway intrusion scenarios.

Datasets related to railway environments---CES, RailFOD23 \cite{Chen2024RailFOD23}, RailGen, and real RFOD scenes---achieve significantly higher scores, demonstrating that the evaluation protocol aligns well with the task semantics. Our method achieves an average score of 5.21, substantially outperforming RailFOD23 (3.54) and approaching real-scene performance. This indicates the high fidelity and applicability of the generated samples. Although real RFOD scenes obtain the highest average score (6.84), their SR score is slightly lower than that of RailGen. We attribute this to the presence of complex noise, lighting variation, and imperfect imaging conditions in real-world data, whereas generated samples tend to be cleaner and more structurally regular, leading to higher perceived scene realism under the evaluation protocol.

\begin{table*}[htbp!]
    \centering
    \caption{Performance comparison with SOTA detection models.}
    \label{tab:performance_comparison_updated}
    \setlength{\tabcolsep}{6pt}
    \begin{tabular}{l | c | cc | cc | cc}
        \toprule
        Model & Params & \multicolumn{2}{c|}{Real Val. Set} & \multicolumn{2}{c|}{RailGen} & \multicolumn{2}{c}{Improvement ($\Delta$)} \\
        \cmidrule(lr){3-4} \cmidrule(lr){5-6} \cmidrule(lr){7-8}
         & & $mAP_{50}$ & $mAP_{50-95}$ & $mAP_{50}$ & $mAP_{50-95}$ & $\Delta_{50}$ & $\Delta_{50-95}$ \\
        \midrule
        YOLO-Master \cite{lin2026yolomaster} & \textbf{9.7M} & 52.10 & 28.10 & 52.70 & 27.00 & $+$0.60 & $-$1.10 \\
        YOLO-World \cite{Cheng2024YOLOWorld}  & 12.8M & 68.80 & 41.60 & 71.10 & 42.60 & $+$2.30 & $+$1.00 \\
        DETR \cite{carion2020detr}           & 41.0M & 62.00 & 35.00 & 65.50 & 39.20 & $+$3.50 & $+$4.20 \\
        Deformable DETR \cite{DBLP:journals/corr/abs-2010-04159}                    & 40.0M & 66.80 & 37.50 & 70.10 & 42.10 & $+$3.30 & $+$4.60 \\
        DINO \cite{caron2021emerging}                                & 47.0M & 69.20 & 41.20 & 72.50 & 44.80 & $+$3.30 & $+$3.60 \\
        Co-DETR \cite{zong2023detrs}                             & 52.1M & 68.50 & 40.80 & 73.20 & 45.20 & $+$4.70 & $+$4.40 \\
        DEIM \cite{huang2025deim}                                & 10.0M & 68.40 & 38.90 & 72.70 & 44.70 & $+$4.30 & $+$5.80 \\
        DEIMv2 \cite{huang2025deimv2}                              & 10.3M & 64.50 & 37.09 & 73.40 & 44.30 & \textbf{+8.90} & \textbf{+7.21} \\
        \textbf{FocalDEIM (Ours)}            & 12.6M & \textbf{71.50} & \textbf{43.50} & \textbf{74.00} & \textbf{46.40} & $+$2.50 & $+$2.90 \\
        \bottomrule
    \end{tabular}
    \\[0.6ex]
    \centering
    \scriptsize\textit{Note: Improvement ($\Delta$) is calculated as the performance on RailGen minus that on Real Val Set (i.e., $\Delta = \text{RailGen} - \text{Real Val Set}$).}
\end{table*}

\subsection{Evaluation of Inter-Class Boundary Clarification via Detection}

As shown in Table~\ref{tab:performance_comparison_updated}, FocalDEIM achieves the best performance under comparable parameter scales (10.0M--12.6M), outperforming the DEIM baseline by 3.10 points in $mAP_{50}$ (71.50 vs.\ 68.40). This validates that focal modulation enhances feature discriminability and improves localization precision for small foreign objects in complex railway scenes.

Most models show performance improvement after incorporating RailGen data, proving its effectiveness in alleviating feature space incompleteness. However, the gains vary across architectures. Notably, DINO achieves 69.20 $mAP_{50}$ on the Real Val Set, making it one of the strongest detectors. Under the RailGen-augmented setting, FocalDEIM reaches 74.00 $mAP_{50}$, surpassing DINO by 4.8 points. This cross-setting comparison highlights the effectiveness of combining feature space expansion with focal-driven discriminative learning.

While DEIMv2 shows the largest numerical gains after augmentation ($\Delta_{50} = 8.90$), its baseline performance remains lower, suggesting greater sensitivity to feature sparsity. In contrast, FocalDEIM maintains the highest absolute accuracy across both real and augmented settings, indicating that focal modulation provides a more robust and discriminative representation under domain shift. Meanwhile, YOLO-Master shows limited or even negative gains, which may be attributed to its reliance on fixed grid-based representations that are less adaptable to distribution shifts introduced by synthetic data.

\subsection{Ablation Study on Detection Model}

\begin{table}[htbp!]
    \centering
    \caption{Ablation Study on Detection Model}
    \label{tab:ablation_study}
    \begin{tabular}{cccccc}
        \toprule
        DEIM & FocalBlock & FocalLoss & RailGen & $mAP_{50}$ & $mAP_{50-95}$ \\
        \midrule
        \checkmark &            &            &            & 68.40 & 38.90 \\
        \checkmark & \checkmark &            &            & 70.80 & 40.70 \\
        \checkmark &            & \checkmark &            & 72.20 & 41.50 \\
        \checkmark & \checkmark & \checkmark &            & 71.50 & 43.50 \\
        \midrule
        \checkmark &            &            & \checkmark & 72.70 & 44.70 \\
        \checkmark & \checkmark &            & \checkmark & 72.80 & 46.20 \\
        \checkmark &            & \checkmark & \checkmark & 71.20 & 45.90 \\
        \checkmark & \checkmark & \checkmark & \checkmark & \textbf{74.00} & \textbf{46.40} \\
        \bottomrule
    \end{tabular}
\end{table}

To validate the complementary roles of feature space completion (via RailGen) and inter-class boundary discrimination (via FocalDEIM), we conduct systematic ablation experiments under identical settings. As shown in Table~\ref{tab:ablation_study}, the DEIM baseline achieves 68.40 $mAP_{50}$ and 38.90 $mAP_{50-95}$. Introducing focal components alone consistently improves performance: DEIM + FocalBlock reaches 70.80/40.70, DEIM + FocalLoss reaches 72.20/41.50, and combining both yields 71.50/43.50. These results confirm that each focal component contributes to boundary clarification and discriminative learning.

Incorporating RailGen further boosts performance in most configurations. For example, the DEIM baseline improves from 68.40/38.90 to 72.70/44.70 after adding RailGen. The full model achieves the best result of 74.00/46.40, corresponding to gains of +5.6 $mAP_{50}$ and +7.5 $mAP_{50-95}$ over the baseline. Notably, under the DEIM + FocalLoss setting, $mAP_{50}$ decreases slightly from 72.20 to 71.20 after incorporating RailGen, while $mAP_{50-95}$ increases from 41.50 to 45.90. This suggests that RailGen mainly improves localization quality under stricter IoU thresholds rather than coarse matching at $IoU=0.5$. Overall, these results demonstrate that feature space completion and focal-driven discrimination are complementary, and their joint optimization leads to the strongest detection performance.

\subsection{Sensitivity Analysis of Focal Cost for Boundary Clarification in Hungarian Matching}

\begin{figure}[htbp]
    \centering
    \includegraphics[width=0.98\linewidth]{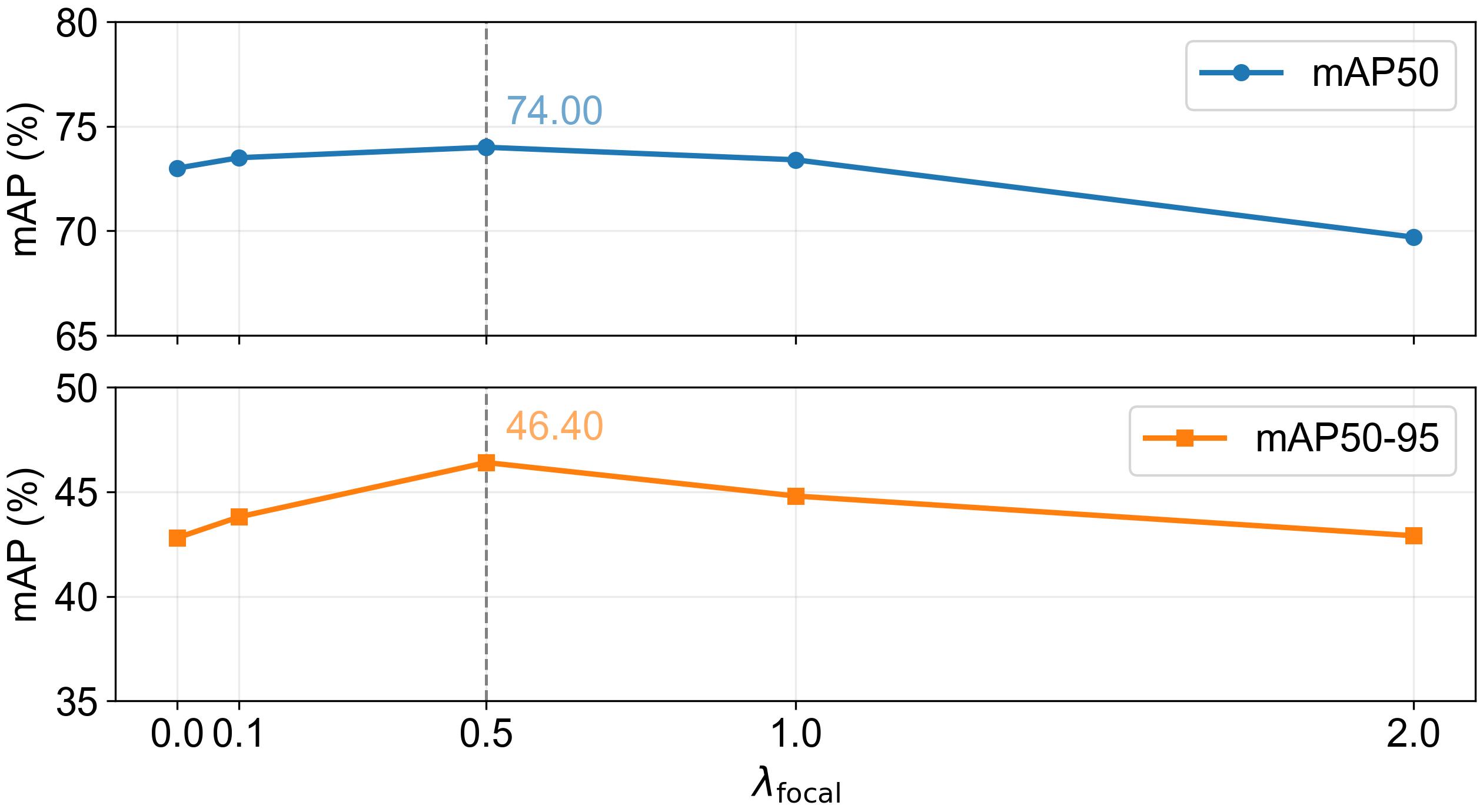}
    \caption{Sensitivity of detection performance to different $\lambda_{\text{focal}}$ values in Hungarian matching.}
    \label{fig:focal_lambda_sensitivity_main}
\end{figure}

To verify the sensitivity of the focal cost weight $\lambda_{\text{focal}}$ in Hungarian matching, we conduct a grid search over $\{0.1, 0.5, 1.0, 2.0\}$. As shown in Fig.~\ref{fig:focal_lambda_sensitivity_main}, the optimal balance is achieved at $\lambda_{\text{focal}}=0.5$. When weights are too small ($<=0.1$), small objects fail to be distinguished from background clutter; when weights are too large ($>=1.0$), the matching process tends to overemphasize feature similarity at the expense of spatial localization accuracy.

Specifically, $\lambda_{\text{focal}}$ balances the focal-aware cost with the original classification and localization terms in the Hungarian matching objective. Based on the empirical results, we adopt $\lambda_{\text{focal}}=0.5$ in all experiments.

\subsection{Comparative Analysis of Feature Space Expansion Strategies}

\begin{table}[htbp]
    \centering
    \caption{Augmentation comparison of different datasets.}
    \label{tab:dataset_augmentation}
    \setlength{\tabcolsep}{8pt}
    \begin{tabular}{l | cc | cc}
        \toprule
        Augmentation & \multicolumn{2}{c|}{Real Test Performance} & \multicolumn{2}{c}{Improvement} \\
        \cmidrule(lr){2-3} \cmidrule(lr){4-5}
        & $mAP_{50}$ & $mAP_{50-95}$ & $\Delta_{50}$ & $\Delta_{50-95}$ \\
        \midrule
        None (Real only)                   & 71.50          & 43.50          & --             & --             \\
        RailFOD23~\cite{Chen2024RailFOD23} & 71.26          & 46.01          & $-$0.24        & $+$2.51        \\
        CES   & 71.92          & 45.15          & $+$0.42        & $+$1.65        \\
        \textbf{RailGen (Ours)}            & \textbf{74.00} & \textbf{46.40} & \textbf{+2.50} & \textbf{+2.90} \\
        \bottomrule
    \end{tabular}
    \\[0.6ex]
    \centering
    \scriptsize\textit{Note: Improvements ($\Delta$) are calculated relative to the ``None (Real only)'' baseline.}
\end{table}

To evaluate the effectiveness of different data sources for feature space expansion, we train the detector with a mix of augmented data and real training data, then evaluate the detector on the Real Val Set. As shown in Table~\ref{tab:dataset_augmentation}, RailGen achieves the best overall results, improving $mAP_{50}$ and $mAP_{50-95}$ by 2.50 and 2.90 points respectively over the real-only baseline. This demonstrates that RailGen is the most effective data source for completing the feature space.

CES also brings consistent improvement on both metrics, although the gains are relatively modest. This suggests that data from similar infrastructure scenarios can provide useful complementary information, but its contribution remains constrained by the lack of precise semantic alignment with railway foreign object intrusion. RailFOD23 shows a different trend: its $mAP_{50}$ decreases slightly by 0.24 points, while its $mAP_{50-95}$ improves by 2.51 points. This indicates that RailFOD23 contributes more to localization quality under stricter IoU thresholds than to coarse detection at $IoU=0.5$. Overall, these results confirm that semantically aligned synthetic data is more effective for feature space expansion than directly introducing existing datasets or data from related scenarios.

\section{Conclusion}

This paper proposes a generative-augmented detection paradigm for long-tailed small-object detection. By coupling RailGen and FocalDEIM, the framework jointly addresses data scarcity and feature ambiguity under long-tailed distributions: RailGen enriches the feature space of rare, small objects through semantically consistent multimodal generation, while FocalDEIM sharpens discrimination via Focal Modulation and Focal Loss. This joint design improves sample diversity and feature separability in complex railway scenes. Experimental validation on real-world high-speed railway datasets shows that the proposed multimodal framework consistently outperforms the DEIM baseline and achieves superior performance on rare intrusion patterns and hard examples.

However, generated samples may still exhibit structural discontinuities and texture misalignment, indicating insufficient high-level semantic constraints in multimodal generation. Future work will evolve the agent toward online training guidance with real-time detection feedback, incorporate structure-aware control mechanisms (e.g., Canny-based conditioning), and explore tighter integration of multimodal large models with detectors. The successful validation in railway foreign object detection lays the foundation for extending this multimodal generative-augmented paradigm to broader long-tailed small-object, few-shot, and safety-critical multimedia detection tasks.

\bibliographystyle{IEEEtran}
\bibliography{IEEEabrv,references} 

@STRING{NIPS        = "Proc. Adv. Neural Inf. Process. Syst."}

@STRING{ICCV        = "Proc. {IEEE} Int. Conf. Comput. Vis."}

@STRING{ICLR        = "Proc. Int. Conf. Learn. Representations"}

@STRING{ECCV        = "Proc. Eur. Conf. Comput. Vis."}

@STRING{CVPR        = "Proc. {IEEE/CVF} Conf. Comput. Vis. Pattern Recognit."}

@STRING{IJCV        = "Int. J. Comput. Vis."}

@STRING{IEEE_J_ITS        = "{IEEE} Trans. Intell. Transp. Syst."}

@STRING{IEEE_J_IM         = "{IEEE} Trans. Instrum. Meas."}

@STRING{SCIDATA      = "Sci. Data"}

@STRING{SCIREP       = "Sci. Rep."}

@STRING{SENSORS      = "Sensors"}

@STRING{JBIGDATA     = "J. Big Data"}

@STRING{INNOVATION   = "Innovation"}

@STRING{APPSCI       = "Appl. Sci."}

@STRING{AISTATS = "Int. Conf. Artif. Intell. Stat."}

@inproceedings{caron2021emerging,
  title={Emerging Properties in Self-Supervised Vision Transformers},
  author={Caron, Mathilde and Touvron, Hugo and Misra, Ishan and J\'egou, Herv\'e  and Mairal, Julien and Bojanowski, Piotr and Joulin, Armand},
  booktitle=ICCV,
  year={2021}
}

@inproceedings{zong2023detrs,
  title={Detrs with collaborative hybrid assignments training},
  author={Zong, Zhuofan and Song, Guanglu and Liu, Yu},
  booktitle=ICCV,
  pages={6748--6758},
  year={2023}
}

@misc{nanobanana,
  author       = {{Google DeepMind}},
  title        = {{Nano Banana}},
  howpublished = {\url{https://deepmind.google/models/gemini-image/}},
}

@article{DBLP:journals/corr/abs-2010-04159,
  author       = {Xizhou Zhu and
                  Weijie Su and
                  Lewei Lu and
                  Bin Li and
                  Xiaogang Wang and
                  Jifeng Dai},
  title        = {Deformable {DETR:} Deformable Transformers for End-to-End Object Detection},
  journal      = {CoRR},
  volume       = {abs/2010.04159},
  year         = {2020}
}

@misc{zhang2023adding,
  title={Adding Conditional Control to Text-to-Image Diffusion Models}, 
  author={Lvmin Zhang and Anyi Rao and Maneesh Agrawala},
  booktitle={ICCV},
  year={2023},
}

@misc{han2021soda10m,
      title={SODA10M: A Large-Scale 2D Self/Semi-Supervised Object Detection Dataset for Autonomous Driving}, 
      author={Jianhua Han and Xiwen Liang and Hang Xu and Kai Chen and Lanqing Hong and Jiageng Mao and Chaoqiang Ye and Wei Zhang and Zhenguo Li and Xiaodan Liang and Chunjing Xu},
      year={2021},
      eprint={2106.11118},
      archivePrefix={arXiv},
      primaryClass={cs.CV}
}

@inproceedings{coates2011analysis,
  author    = {Adam Coates and Honglak Lee and Andrew Y. Ng},
  title     = {An Analysis of Single-Layer Networks in Unsupervised Feature Learning},
  booktitle = AISTATS,
  pages     = {215--223},
  year      = {2011}
}

@article{Chen2024RailFOD23,
  author    = {Chen, Zhichao and Yang, Jie and Feng, Zhicheng and Zhu, Hao},
  title     = {RailFOD23: A dataset for foreign object detection on railroad transmission lines},
  journal   = SCIDATA,
  year      = {2024},
  volume    = {11},
  number    = {1},
  pages     = {72},
  doi       = {10.1038/s41597-024-02918-9},
  issn      = {2052-4463},
  publisher = {Nature Publishing Group},
  month     = jan,
  day       = {16}
}

@misc{gemmateam2025gemma3technicalreport,
  title         = {Gemma 3 Technical Report},
  author        = {Gemma Team},
  year          = {2025},
  eprint        = {2503.19786},
  archivePrefix = {arXiv},
  primaryClass  = {cs.CL},
  url           = {https://arxiv.org/abs/2503.19786}
}

@misc{lipman2023flowmatchinggenerativemodeling,
  title         = {Flow Matching for Generative Modeling},
  author        = {Yaron Lipman and Ricky T. Q. Chen and Heli Ben-Hamu and Maximilian Nickel and Matt Le},
  year          = {2023},
  eprint        = {2210.02747},
  archivePrefix = {arXiv},
  primaryClass  = {cs.LG},
  url           = {https://arxiv.org/abs/2210.02747}
}

@article{pei2025neuro,
  title={Neuro-vae-symbolic dynamic traffic management},
  author={Pei, Jiaming and Li, Jinhai and Song, Zhenyu and Dabel, Maryam Mohamed Al and Alenazi, Mohammed J. F. and Zhang, Sun and Bashir, Ali Kashif},
  journal= IEEE_J_ITS ,
  year={2025},
  publisher={IEEE}
}

@misc{goodfellow2014generativeadversarialnetworks,
  title         = {Generative Adversarial Networks},
  author        = {Ian J. Goodfellow and Jean Pouget-Abadie and Mehdi Mirza and Bing Xu and David Warde-Farley and Sherjil Ozair and Aaron Courville and Yoshua Bengio},
  year          = {2014},
  eprint        = {1406.2661},
  archivePrefix = {arXiv},
  primaryClass  = {stat.ML},
  url           = {https://arxiv.org/abs/1406.2661}
}

@ARTICLE{10931847,
  author={Kou, Feifei and Yao, Yuhan and Han, Jideng and Wang, Jiahao and Li, Haisheng and Li, Xingchen and Zhang, Jiwei},
  journal=IEEE_J_ITS, 
  title={DualFocus GAN for Robust Watermarking in Transportation Cyber-Physical Systems}, 
  year={2025},
  volume={26},
  number={9},
  pages={14371-14382},
  doi={10.1109/TITS.2025.3550120}}

@inproceedings{hu2022lora,
  title     = {Lo{RA}: Low-Rank Adaptation of Large Language Models},
  author    = {Edward J Hu and Yelong Shen and Phillip Wallis and Zeyuan Allen-Zhu and Yuanzhi Li and Shean Wang and Lu Wang and Weizhu Chen},
  booktitle = ICLR,
  year      = {2022},
  url       = {https://openreview.net/forum?id=nZeVKeeFYf9}
}

@article{GU2026101253,
title = {A survey on LLM-as-a-judge},
journal = {The Innovation},
pages = {101253},
year = {2026},
issn = {2666-6758},
doi = {https://doi.org/10.1016/j.xinn.2025.101253},
url = {https://www.sciencedirect.com/science/article/pii/S2666675825004564},
author = {Jiawei Gu and Xuhui Jiang and Zhichao Shi and Hexiang Tan and Xuehao Zhai and Chengjin Xu and Wei Li and Yinghan Shen and Shengjie Ma and Honghao Liu and Saizhuo Wang and Kun Zhang and Zhouchi Lin and Bowen Zhang and Lionel Ni and Wen Gao and Yuanzhuo Wang and Jian Guo},
}

@misc{labs2025flux1kontextflowmatching,
  title         = {FLUX.1 Kontext: Flow Matching for In-Context Image Generation and Editing in Latent Space},
  author        = {Black Forest Labs and Stephen Batifol and Andreas Blattmann and Frederic Boesel and Saksham Consul and Cyril Diagne and Tim Dockhorn and Jack English and Zion English and Patrick Esser and Sumith Kulal and Kyle Lacey and Yam Levi and Cheng Li and Dominik Lorenz and Jonas Müller and Dustin Podell and Robin Rombach and Harry Saini and Axel Sauer and Luke Smith},
  year          = {2025},
  eprint        = {2506.15742},
  archivePrefix = {arXiv},
  primaryClass  = {cs.GR},
  url           = {https://arxiv.org/abs/2506.15742}
}

@misc{flux2024,
  author       = {Black Forest Labs},
  title        = {FLUX},
  year         = {2024},
  howpublished = {\url{https://github.com/black-forest-labs/flux}}
}

@article{tian2025yolov12,
  title   = {Yolov12: Attention-centric real-time object detectors},
  author  = {Tian, Yunjie and Ye, Qixiang and Doermann, David},
  journal = {arXiv preprint arXiv:2502.12524},
  year    = {2025}
}

@article{lei2025yolov13,
  title   = {YOLOv13: Real-Time Object Detection with Hypergraph-Enhanced Adaptive Visual Perception},
  author  = {Lei, Mengqi and Li, Siqi and Wu, Yihong and Hu, Han and Zhou, You and Zheng, Xinhu and Ding, Guiguang and Du, Shaoyi and Wu, Zongze and Gao, Yue},
  journal = {arXiv preprint arXiv:2506.17733},
  year    = {2025}
}

@article{yang2022focal,
  title   = {Focal modulation networks},
  author  = {Yang, Jianwei and Li, Chunyuan and Dai, Xiyang and Gao, Jianfeng},
  journal = NIPS,
  volume  = {35},
  pages   = {4203--4217},
  year    = {2022}
}

@article{cui2026focal,
  title     = {Focal Modulation for Image Restoration: Y. Cui et al.},
  author    = {Cui, Yuning and Ren, Wenqi and Knoll, Alois},
  journal   = IJCV,
  volume    = {134},
  number    = {1},
  pages     = {6},
  year      = {2026},
  publisher = {Springer}
}

@inproceedings{kirillov2023segment,
  title     = {Segment anything},
  author    = {Kirillov, Alexander and Mintun, Eric and Ravi, Nikhila and Mao, Hanzi and Rolland, Chloe and Gustafson, Laura and Xiao, Tete and Whitehead, Spencer and Berg, Alexander C and Lo, Wan-Yen and others},
  booktitle = ICCV,
  pages     = {4015--4026},
  year      = {2023}
}

@InProceedings{lueddecke22_cvpr,
  author    = {L\"uddecke, Timo and Ecker, Alexander},
  title     = {Image Segmentation Using Text and Image Prompts},
  booktitle = CVPR,
  month     = jun,
  year      = {2022},
  pages     = {7086--7096}
}

@misc{kingma2022autoencodingvariationalbayes,
  title         = {Auto-Encoding Variational Bayes},
  author        = {Diederik P Kingma and Max Welling},
  year          = {2022},
  eprint        = {1312.6114},
  archivePrefix = {arXiv},
  primaryClass  = {stat.ML},
  url           = {https://arxiv.org/abs/1312.6114}
}

@misc{radford2021learningtransferablevisualmodels,
  title         = {Learning Transferable Visual Models From Natural Language Supervision},
  author        = {Alec Radford and Jong Wook Kim and Chris Hallacy and Aditya Ramesh and Gabriel Goh and Sandhini Agarwal and Girish Sastry and Amanda Askell and Pamela Mishkin and Jack Clark and Gretchen Krueger and Ilya Sutskever},
  year          = {2021},
  eprint        = {2103.00020},
  archivePrefix = {arXiv},
  primaryClass  = {cs.CV},
  url           = {https://arxiv.org/abs/2103.00020}
}

@article{zhou2026semantic,
  title={Semantic Image Synthesis via Diffusion Models},
  author={Zhou, Wengang and Wang, Weilun and Bao, Jianmin and Chen, Dongdong and Chen, Dong and Yuan, Lu and Li, Houqiang},
  journal={IEEE Transactions on Multimedia},
  year={2026},
  publisher={IEEE}
}

@inproceedings{huang2025deim,
  title     = {Deim: Detr with improved matching for fast convergence},
  author    = {Huang, Shihua and Lu, Zhichao and Cun, Xiaodong and Yu, Yongjun and Zhou, Xiao and Shen, Xi},
  booktitle = CVPR,
  pages     = {15162--15171},
  year      = {2025}
}

@article{huang2025deimv2,
  title={Real-Time Object Detection Meets DINOv3},
  author={Huang, Shihua and Hou, Yongjie and Liu, Longfei and Yu, Xuanlong and Shen, Xi},
  journal={arXiv},
  year={2025}
}

@inproceedings{zhao2024detrs,
  title     = {Detrs beat yolos on real-time object detection},
  author    = {Zhao, Yian and Lv, Wenyu and Xu, Shangliang and Wei, Jinman and Wang, Guanzhong and Dang, Qingqing and Liu, Yi and Chen, Jie},
  booktitle = CVPR,
  pages     = {16965--16974},
  year      = {2024}
}

@article{shehzadi2025object,
  title     = {Object detection with transformers: A review},
  author    = {Shehzadi, Tahira and Hashmi, Khurram Azeem and Liwicki, Marcus and Stricker, Didier and Afzal, Muhammad Zeshan},
  journal   = SENSORS,
  volume    = {25},
  number    = {19},
  pages     = {6025},
  year      = {2025},
  publisher = {MDPI}
}

@article{dai2023adaptive,
  title={An adaptive sample assignment network for tiny object detection},
  author={Dai, Honghao and Gao, Shanshan and Huang, Hong and Mao, Deqian and Zhang, Chenhao and Zhou, Yuanfeng},
  journal={IEEE Transactions on Multimedia},
  volume={26},
  pages={2918--2931},
  year={2023},
  publisher={IEEE}
}

@article{deng2021extended,
  title={Extended feature pyramid network for small object detection},
  author={Deng, Chunfang and Wang, Mengmeng and Liu, Liang and Liu, Yong and Jiang, Yunliang},
  journal={IEEE Transactions on Multimedia},
  volume={24},
  pages={1968--1979},
  year={2021},
  publisher={IEEE}
}

@article{yin2024sdpdet,
  title={SDPDet: Learning scale-separated dynamic proposals for end-to-end drone-view detection},
  author={Yin, Nengzhong and Liu, Chengxu and Tian, Ruhao and Qian, Xueming},
  journal={IEEE transactions on multimedia},
  volume={26},
  pages={7812--7822},
  year={2024},
  publisher={IEEE}
}

@article{yin2025foodlmm,
  title={Foodlmm: A versatile food assistant using large multi-modal model},
  author={Yin, Yuehao and Qi, Huiyan and Zhu, Bin and Chen, Jingjing and Jiang, Yu-Gang and Ngo, Chong-Wah},
  journal={IEEE Transactions on Multimedia},
  year={2025},
  publisher={IEEE}
}

@article{wang2025detailed,
  title={Detailed Object Description With Controllable Dimensions},
  author={Wang, Xinran and Zhang, Haiwen and Li, Baoteng and Liang, Kongming and Sun, Hao and He, Zhongjiang and Ma, Zhanyu and Guo, Jun},
  journal={IEEE Transactions on Multimedia},
  volume={27},
  pages={8474--8485},
  year={2025}
}

@article{gu2024survey,
  title     = {A survey on llm-as-a-judge},
  author    = {Gu, Jiawei and Jiang, Xuhui and Shi, Zhichao and Tan, Hexiang and Zhai, Xuehao and Xu, Chengjin and Li, Wei and Shen, Yinghan and Ma, Shengjie and Liu, Honghao and others},
  journal   = INNOVATION,
  year      = {2024},
  publisher = {Elsevier}
}

@article{li2023yolov5s,
  title     = {YOLOv5s-D: A railway catenary dropper state identification and small defect detection model},
  author    = {Li, Ziyi and Rao, Zhiqiang and Ding, Lu and Ding, Biao and Fang, Jianjun and Ma, Xiaoning},
  journal   = APPSCI,
  volume    = {13},
  number    = {13},
  pages     = {7881},
  year      = {2023},
  publisher = {MDPI}
}

@article{jin2025optimized,
  title     = {Optimized YOLOv11m for real-time high-speed railway catenary defect detection},
  author    = {Jin, Tao and Shen, Zhijun and Geng, Haowen},
  journal   = SCIREP,
  year      = {2025},
  publisher = {Nature Publishing Group UK London}
}

@inproceedings{rombach2022high,
  title     = {High-resolution image synthesis with latent diffusion models},
  author    = {Rombach, Robin and Blattmann, Andreas and Lorenz, Dominik and Esser, Patrick and Ommer, Bj{\"o}rn},
  booktitle = CVPR,
  pages     = {10684--10695},
  year      = {2022}
}

@ARTICLE{10902119,
  author={He, Jin and Wang, Wei and Lv, Fengmao and Luo, Haonan and Zhang, Gexiang and Chen, Zhenghua},
  journal= IEEE_J_ITS, 
  title={Multi-Scale CNN-Transformer Hybrid Network for Rail Fastener Defect Detection}, 
  year={2025},
  volume={26},
  number={6},
  pages={8894-8906},
  doi={10.1109/TITS.2025.3540846}}

@article{alzubaidi2023survey,
  title     = {A survey on deep learning tools dealing with data scarcity: definitions, challenges, solutions, tips, and applications},
  author    = {Alzubaidi, Laith and Bai, Jinshuai and Al-Sabaawi, Aiman and Santamar{\'\i}a, Jose and Albahri, Ahmed Shihab and Al-Dabbagh, Bashar Sami Nayyef and Fadhel, Mohammed A and Manoufali, Mohamed and Zhang, Jinglan and Al-Timemy, Ali H and others},
  journal   = JBIGDATA,
  volume    = {10},
  number    = {1},
  pages     = {46},
  year      = {2023},
  publisher = {Springer}
}

@article{yang2022survey,
  title     = {A survey on long-tailed visual recognition},
  author    = {Yang, Lu and Jiang, He and Song, Qing and Guo, Jun},
  journal   = IJCV,
  volume    = {130},
  number    = {7},
  pages     = {1837--1872},
  year      = {2022},
  publisher = {Springer}
}

@ARTICLE{11175219,
  author   = {Xu, Xin-Yue and Wang, Su-Mei and Liu, Wen-Qiang and Ni, Yi-Qing},
  journal  = IEEE_J_IM,
  title    = {Advancements in Obstacle Intrusion Detection Methods for Rail Transit: A Comprehensive Review},
  year     = {2025},
  volume   = {74},
  number   = {},
  pages    = {1--34},
  doi      = {10.1109/TIM.2025.3612624}
}

@ARTICLE{11237202,
  author   = {Hao, Quan and Shi, Rui and Li, Jiaze and Zhang, Liguo},
  journal  = IEEE_J_ITS,
  title    = {Generative Approach for Detecting Small Intrusive Foreign Objects in High-Speed Railway Scenario},
  year     = {2026},
  volume   = {27},
  number   = {1},
  pages    = {1471--1484},
  doi      = {10.1109/TITS.2025.3625181}
}

@STRING{AEI          = "Adv. Eng. Inform."}

@STRING{MEASURE      = "Measurement"}

@STRING{JIII         = "J. Ind. Inf. Integr."}

@STRING{IEEE_J_MECTH = "{IEEE/ASME} Trans. Mechatronics"}

@ARTICLE{10719679,
  author   = {Song, Xiying and Song, Haifeng and Wang, Hongwei and Zhang, Zixuan and Dong, Hairong},
  journal  = IEEE_J_MECTH,
  title    = {Deep Learning-Based Railway Foreign Object Intrusion Intelligent Perception Using Attention-Aggregated Semantic Segmentation},
  year     = {2025},
  volume   = {30},
  number   = {4},
  pages    = {2609--2619},
  doi      = {10.1109/TMECH.2024.3468620}
}

@article{bai2026rosd,
  title     = {ROSD: Railway intrusion object generalized detection via Open-Set Detection},
  author    = {Bai, Dingyuan and Guo, Baoqing and Yu, Zujun and Ruan, Tao and Zhou, Xingfang and Sun, Tao},
  journal   = AEI,
  volume    = {71},
  pages     = {104228},
  year      = {2026},
  publisher = {Elsevier}
}

@article{cao2022effective,
  title     = {An effective railway intrusion detection method using dynamic intrusion region and lightweight neural network},
  author    = {Cao, Zhiwei and Qin, Yong and Xie, Zhengyu and Liu, Qinghong and Zhang, Ehui and Wu, Zhiyu and Yu, Zujun},
  journal   = MEASURE,
  volume    = {191},
  pages     = {110564},
  year      = {2022},
  publisher = {Elsevier}
}

@article{zhang2024railway,
  title     = {Railway obstacle intrusion warning mechanism integrating YOLO-based detection and risk assessment},
  author    = {Zhang, Zhipeng and Chen, Peiru and Huang, Yujie and Dai, Lei and Xu, Feng and Hu, Hao},
  journal   = JIII,
  volume    = {38},
  pages     = {100571},
  year      = {2024},
  publisher = {Elsevier}
}

@inproceedings{Cheng2024YOLOWorld,
  author    = {Cheng, T. and Song, L. and Ge, Y. and Liu, W. and Wang, X. and Shan, Y.},
  title     = {{YOLO-World}: Real-Time Open-Vocabulary Object Detection},
  booktitle = CVPR,
  pages     = {16901--16911},
  year      = {2024}
}

@inproceedings{lin2026yolomaster,
  author    = {Lin, X. and Peng, J. and Gan, Z. and Zhu, J. and Liu, J.},
  title     = {{YOLO-Master}: {MOE}-Accelerated with Specialized Transformers for Enhanced Real-time Detection},
  booktitle = CVPR,
  year      = {2026}
}

@inproceedings{carion2020detr,
  author    = {Carion, N. and Massa, F. and Synnaeve, G. and Usunier, N. and Kirillov, A. and Zagoruyko, S.},
  title     = {End-to-End Object Detection with Transformers},
  booktitle = {Proc. ECCV},
  year      = {2020}
}

\begin{IEEEbiography}[{\includegraphics[width=1in,height=1.25in,clip,keepaspectratio]{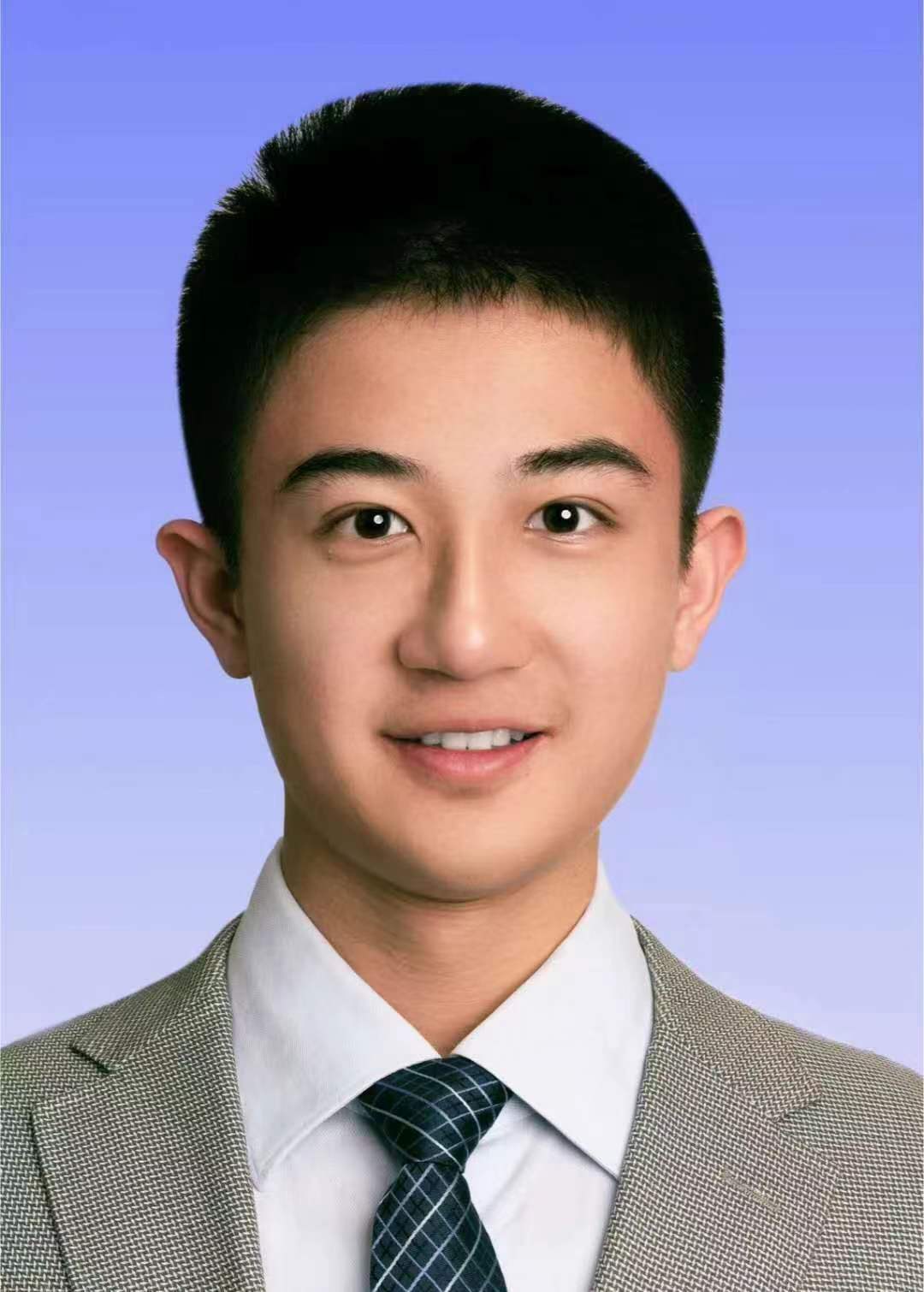}}]{Quan Hao}
    received his bachelor's degree from the College of Software, Beijing University of Technology (BJUT), Beijing, China, in 2022. He is currently a Ph.D. student in the School of Information Science and Technology at the same university. His current research interests include generative artificial intelligence and intelligent transportation system.
\end{IEEEbiography}

\begin{IEEEbiography}[{\includegraphics[width=1in,height=1.25in,clip,keepaspectratio]{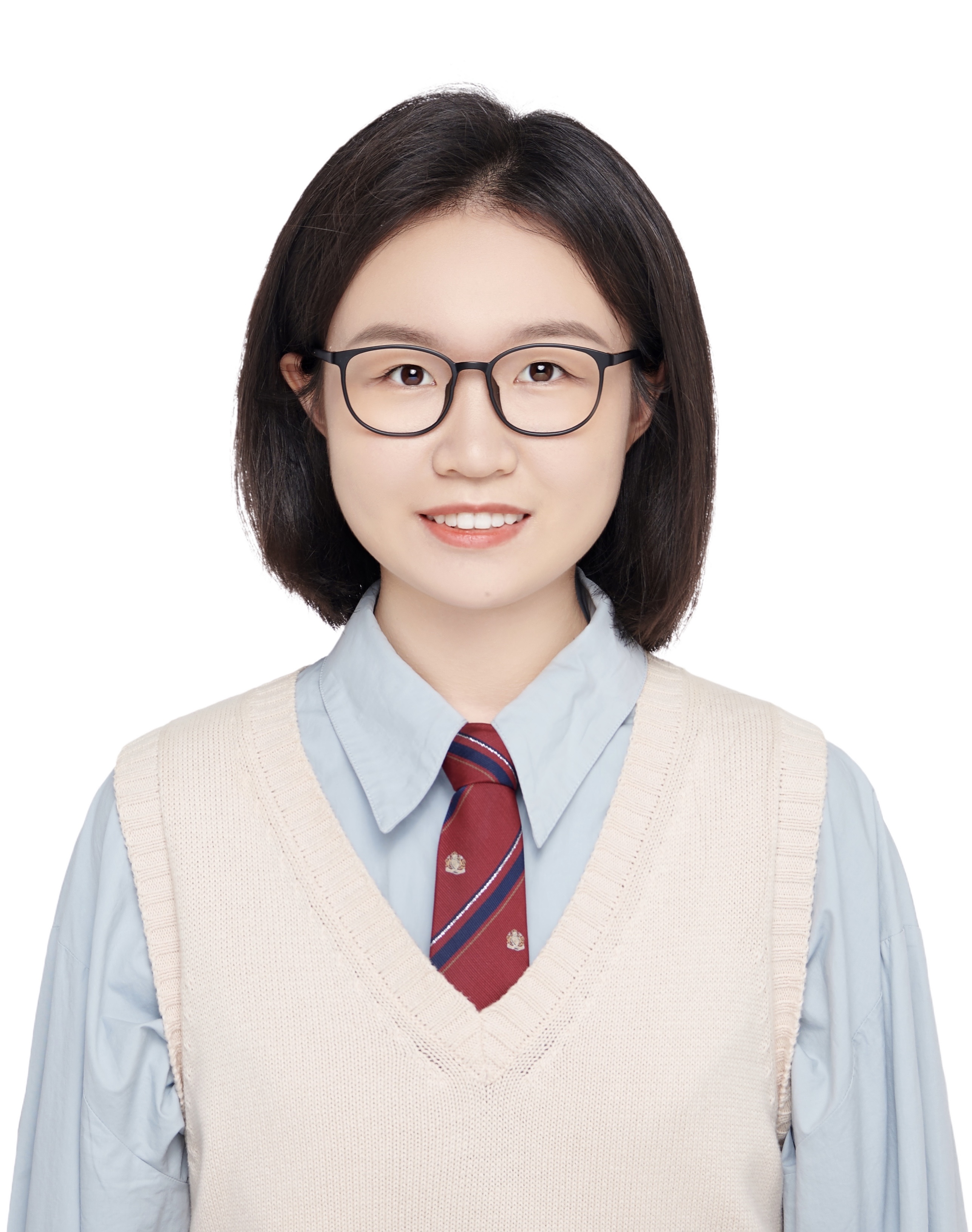}}]{Ziyang Tao}
    is a fourth-year undergraduate student in the College of Computer Science at Beijing University of Technology, where she is pursuing an undergraduate degree in Information Security. Her current research interests include generative artificial intelligence and machine learning.
\end{IEEEbiography}

\begin{IEEEbiography}[{\includegraphics[width=1in,height=1.25in,clip,keepaspectratio]{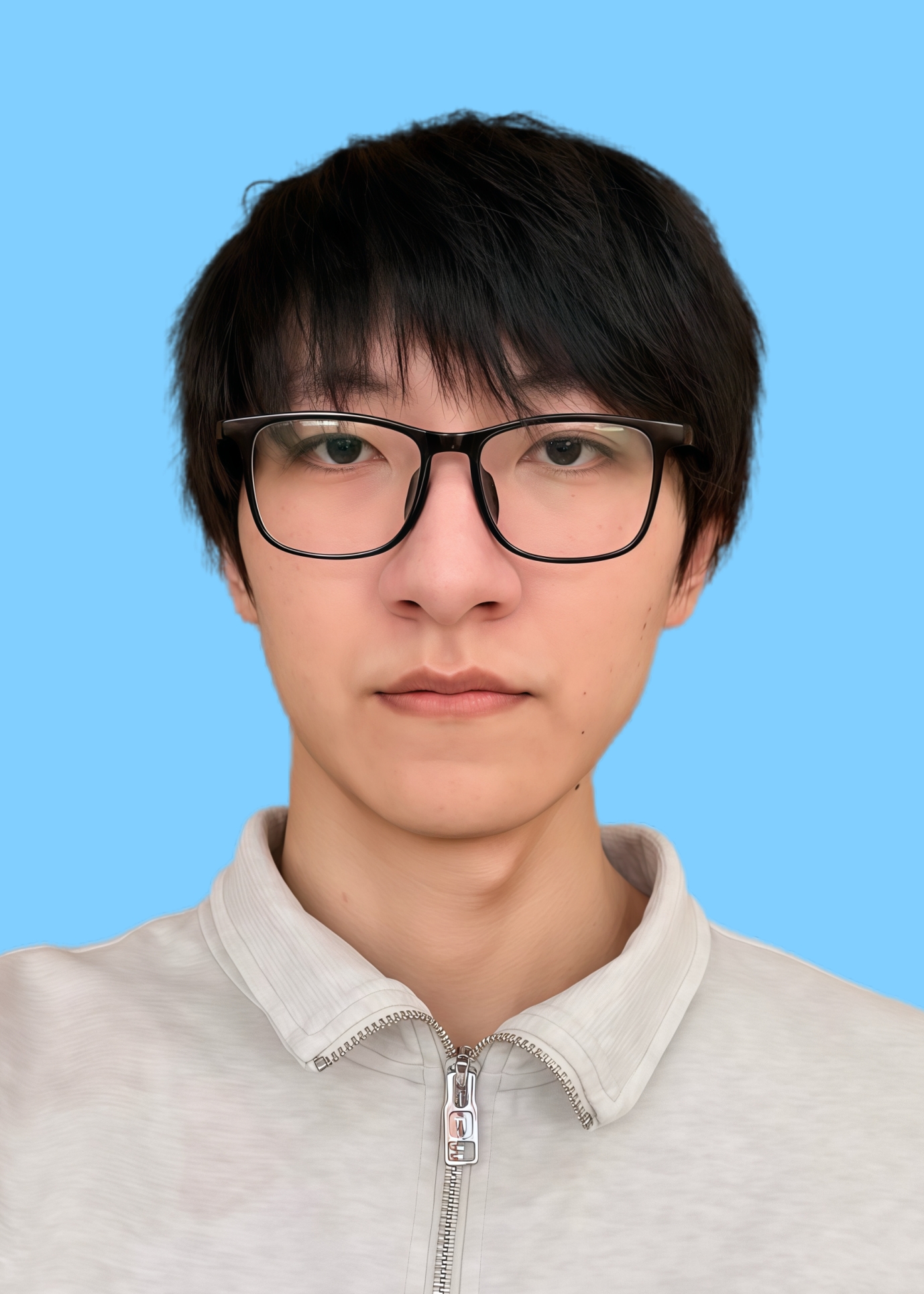}}]{Chenxi Zhang}
    is a third-year undergraduate student in the School of Information Science and Technology at Beijing University of Technology, where he is pursuing an undergraduate degree in Robotics Engineering. His current research interests include autonomous driving and computer vision.
\end{IEEEbiography}

\begin{IEEEbiography}[{\includegraphics[width=1in,height=1.25in,clip,keepaspectratio]{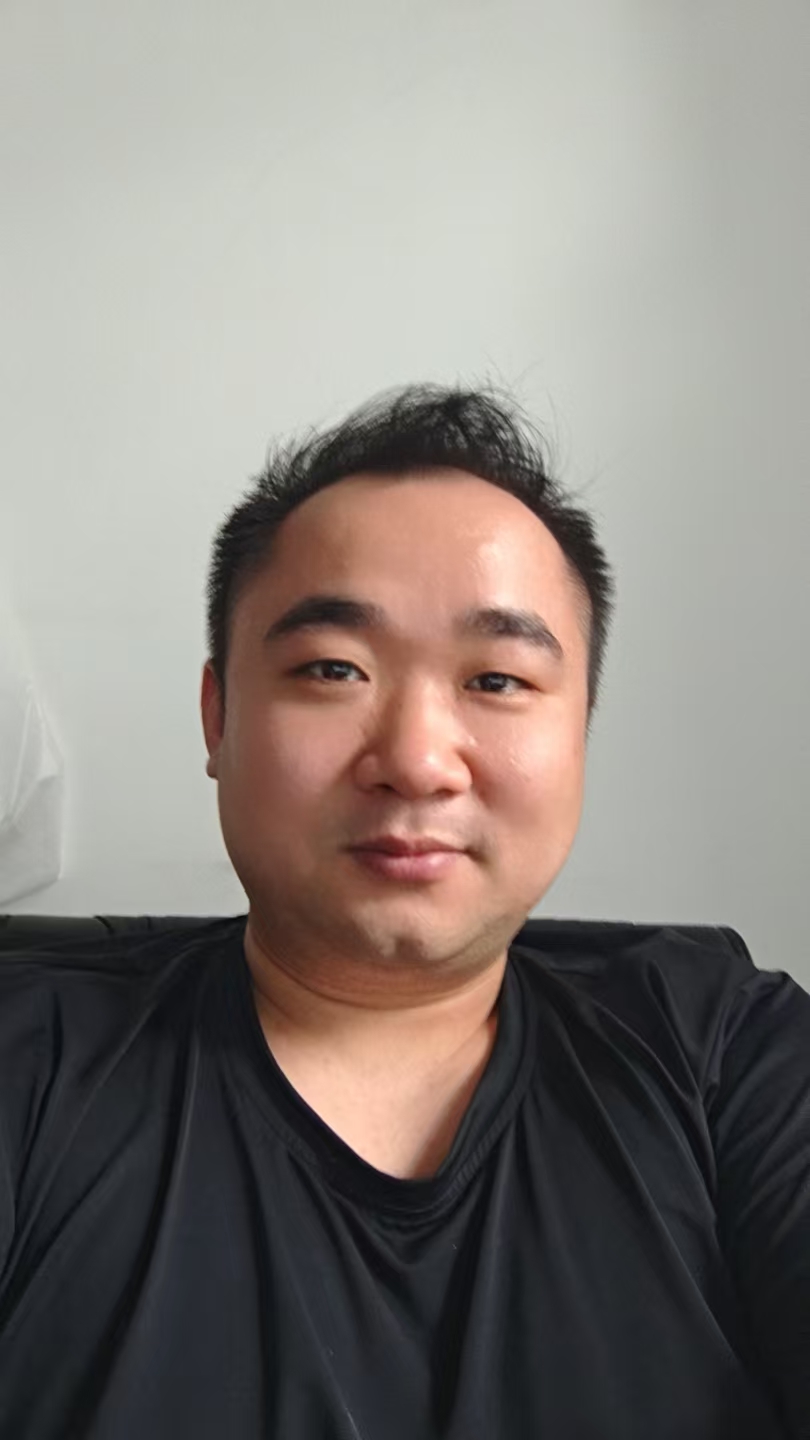}}]{Yudong Wang}
    received his Ph.D. in Computer Science from the Institute of Automation, Chinese Academy of Sciences (CASIA) in 2022. He is currently an assistant professor and lecturer at CASIA. His research interests include intelligent transportation, smart factories, industrial anomaly detection.
\end{IEEEbiography}

\begin{IEEEbiography}[{\includegraphics[width=1in,height=1.25in,clip,keepaspectratio]{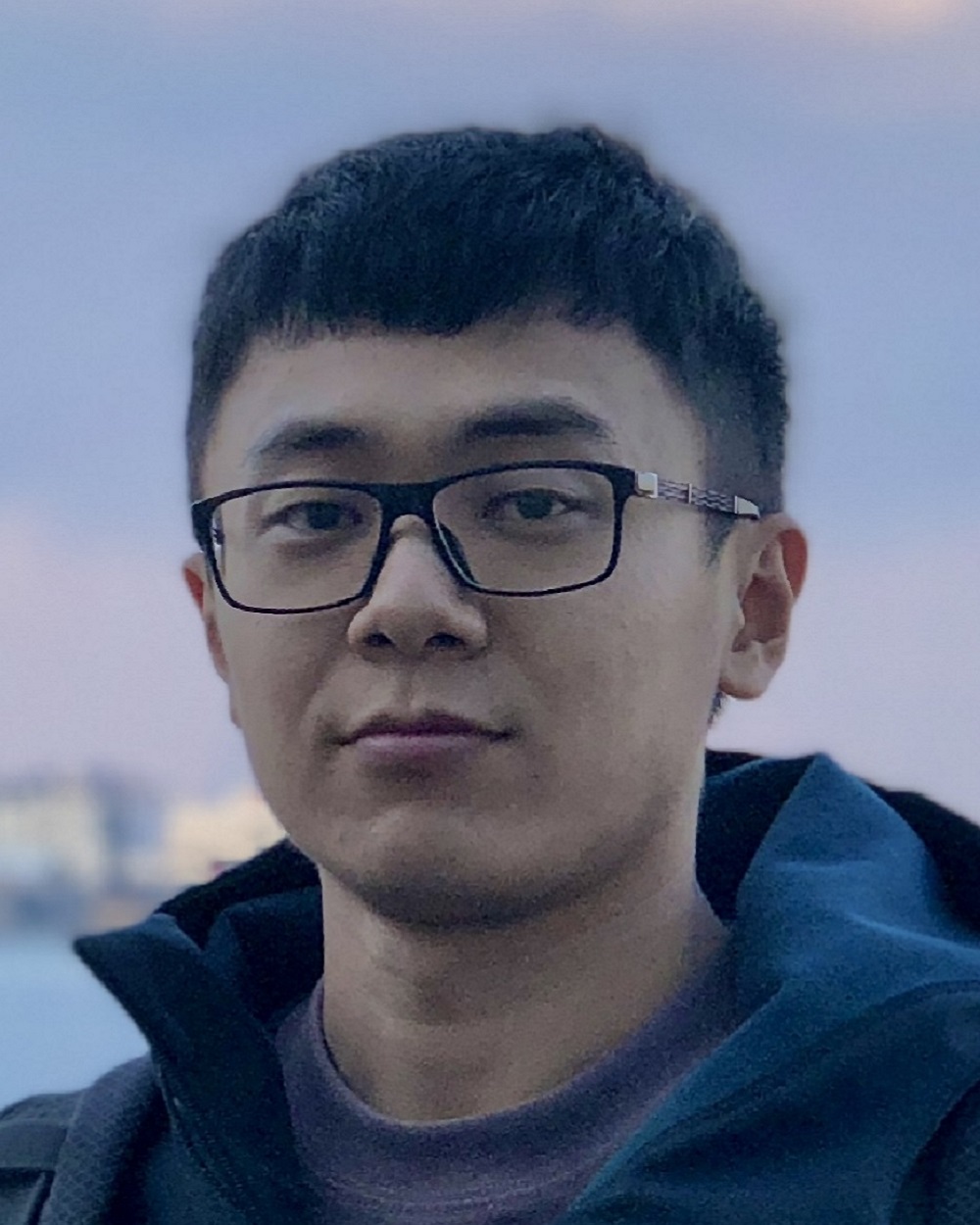}}]{Rui Shi}
    received his Ph.D. degree in graphic and computer sciences from the University of Tokyo, Tokyo, Japan, in 2022. He is currently an associate professor in the School of Information Science and Technology, Beijing University of Technology, Beijing, China. He worked as a visiting researcher with the Department of General Systems Studies, the University of Tokyo. His current research interests include autonomous driving, neural networks, and explainable artificial intelligence.
\end{IEEEbiography}

\begin{IEEEbiography}[{\includegraphics[width=1in,height=1.25in,clip,keepaspectratio]{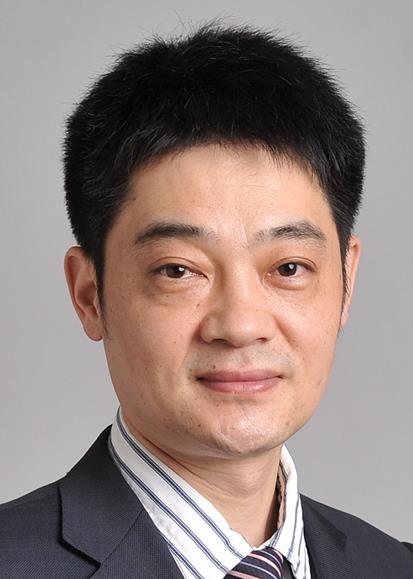}}]{Liguo Zhang (Senior Member)} 
received his Ph.D. degree in control theory and applications from the Beijing University of Technology (BJUT), Beijing, China, in 2006. Since 2014, he has been a Full Professor with the School of Electronic Information and Control Engineering, BJUT. He is currently the Deputy Director of the School of Information Science and Technology, BJUT. His research interests include hybrid systems, intelligent systems, and control of distributed parameter systems. He is an Associate Editor for the IMA Journal Mathematical Control and Information and the Guest Editor of the International Journal of Distributed Sensor Networks.
\end{IEEEbiography}

\vskip 0pt plus -1fil
\vfill


\end{document}